\pdfoutput=1

\documentclass[11pt]{article}

\usepackage[preprint]{acl}

\usepackage{times}
\usepackage{latexsym}

\usepackage[T1]{fontenc}

\usepackage[utf8]{inputenc}

\usepackage{microtype}

\usepackage{inconsolata}

\usepackage{xcolor}
\usepackage{xspace}
\usepackage{amsmath}
\usepackage{graphicx}
\usepackage{booktabs}
\usepackage{array}
\usepackage{makecell}
\usepackage{listings}
\usepackage{subcaption}

\usepackage{multirow}
\usepackage{pbox}

\usepackage{pdfpages}

\usepackage{tikz}
\usetikzlibrary{positioning, fit, shapes}
\usetikzlibrary{bayesnet}
\usetikzlibrary{arrows}

\newcommand{\ie}{\emph{i.e.,}\ }
\newcommand{\eg}{\emph{e.g.,}\ }

\newcommand{\sect}[1]{\S\ref{#1}}
\newcommand{\sectpar}[1]{(\S\ref{#1})}
\newcommand{\abs}[1]{\ensuremath{\left\|#1\right\|}}

\definecolor{lightgray}{RGB}{169, 183, 198} %
\definecolor{darkblue}{RGB}{0, 0, 128} %
\definecolor{darkgray}{RGB}{50, 50, 50} %

\definecolor{lightred}{HTML}{cd5c5c}
\definecolor{green}{HTML}{2ca02c}

\definecolor{gptfour}{HTML}{c94cbe}
\definecolor{random}{HTML}{606264}
\definecolor{ranking}{HTML}{17becf}
\definecolor{gold}{HTML}{1f77b4}
\definecolor{adaptive}{HTML}{9467bd}

\DeclareMathOperator*{\argmax}{\arg\,\max}

\newcommand{\benchmark}{\textsc{AdapT}\xspace}
\newcommand{\benchmarkfull}{\textsc{AdapT} (Adaptive Teaching)\xspace}

\newcommand{\concept}{$h$\xspace}
\newcommand{\conceptmath}{h}
\newcommand{\goldconcept}{$h^*$\xspace}

\newcommand{\fullconceptmath}{\mathcal{H}}

\newcommand{\X}{\ensuremath{\underline{\smash{x}}}\xspace}
\newcommand{\Y}{\ensuremath{\underline{\smash{y}}}\xspace}
\newcommand{\Xspace}{\mathcal{X}}
\newcommand{\Yspace}{\mathcal{Y}}

\newcommand{\gptfour}{\textsc{GPT-4}\xspace}
\newcommand{\gptfourknownteacher}{\textsc{GPT-4-Known}\xspace}
\newcommand{\gptfourunknownteacher}{\textsc{GPT-4}\xspace}
\newcommand{\greedyunknownteacher}{\textsc{Non-Adaptive}\xspace}
\newcommand{\greedyknownteacher}{\textsc{Non-Adaptive-Known}\xspace}
\newcommand{\randomteacher}{\textsc{Random}\xspace}
\newcommand{\rankingknownteacher}{\textsc{Ranking-Known}\xspace}
\newcommand{\rankingunknownteacher}{\textsc{Ranking}\xspace}
\newcommand{\combinedteacher}{\textsc{GPT-4}+\textsc{AToM}\xspace}

\newcommand{\adaptiveunknownteacher}{\textsc{AToM}\xspace}

\newcommand{\undefined}{\emph{undefined}}

\newcommand{\functioninput}{\texttt{x}\xspace}
\newcommand{\wug}{\texttt{wug}\xspace}
\newcommand{\wugx}{\texttt{wug(x)}\xspace}

\newcommand{\f}[1]{\texttt{{#1}}\xspace}

\newcommand{\fknower}{\texttt{b-learner}\xspace}
\newcommand{\gknower}{\texttt{f-learner}\xspace}
\newcommand{\fknowers}{\texttt{b-learner}s\xspace}
\newcommand{\gknowers}{\texttt{f-learner}s\xspace}

\newcommand{\ytoiednotknower}{\texttt{+ied-learner}\xspace}
\newcommand{\dnotknower}{\texttt{+d-learner}\xspace}
\newcommand{\ednotknower}{\texttt{+ed-learner}\xspace}
\newcommand{\consonantednotknower}{\texttt{+consonant+ed-learner}\xspace}

\newcommand{\addgeneralizer}{\texttt{mult-learner}\xspace}
\newcommand{\multgeneralizer}{\texttt{add-learner}\xspace}
\newcommand{\addgeneralizers}{\texttt{mult-learner}s\xspace}
\newcommand{\multgeneralizers}{\texttt{add-learner}s\xspace}

\newcommand{\fnox}{\f{f}}

\newcommand{\fx}{\f{f(x)}}
\newcommand{\linearconcept}{\f{a*x+b}}
\newcommand{\ga}{\f{a}}
\newcommand{\gb}{\f{b}}

\newcommand{\verbclass}[1]{\texttt{{#1}}\xspace}

\newcommand{\codecell}[1]{\ttfamily \footnotesize #1}

\title{Toward In-Context Teaching:\\Adapting Examples to Students' Misconceptions}

\author{Alexis Ross \quad\quad Jacob Andreas \\
        MIT CSAIL \\ 
        {\tt\{alexisro,jda\}@mit.edu}
        }

\begin{document}
\maketitle

\begin{figure}
    \centering \includegraphics[width=\columnwidth]{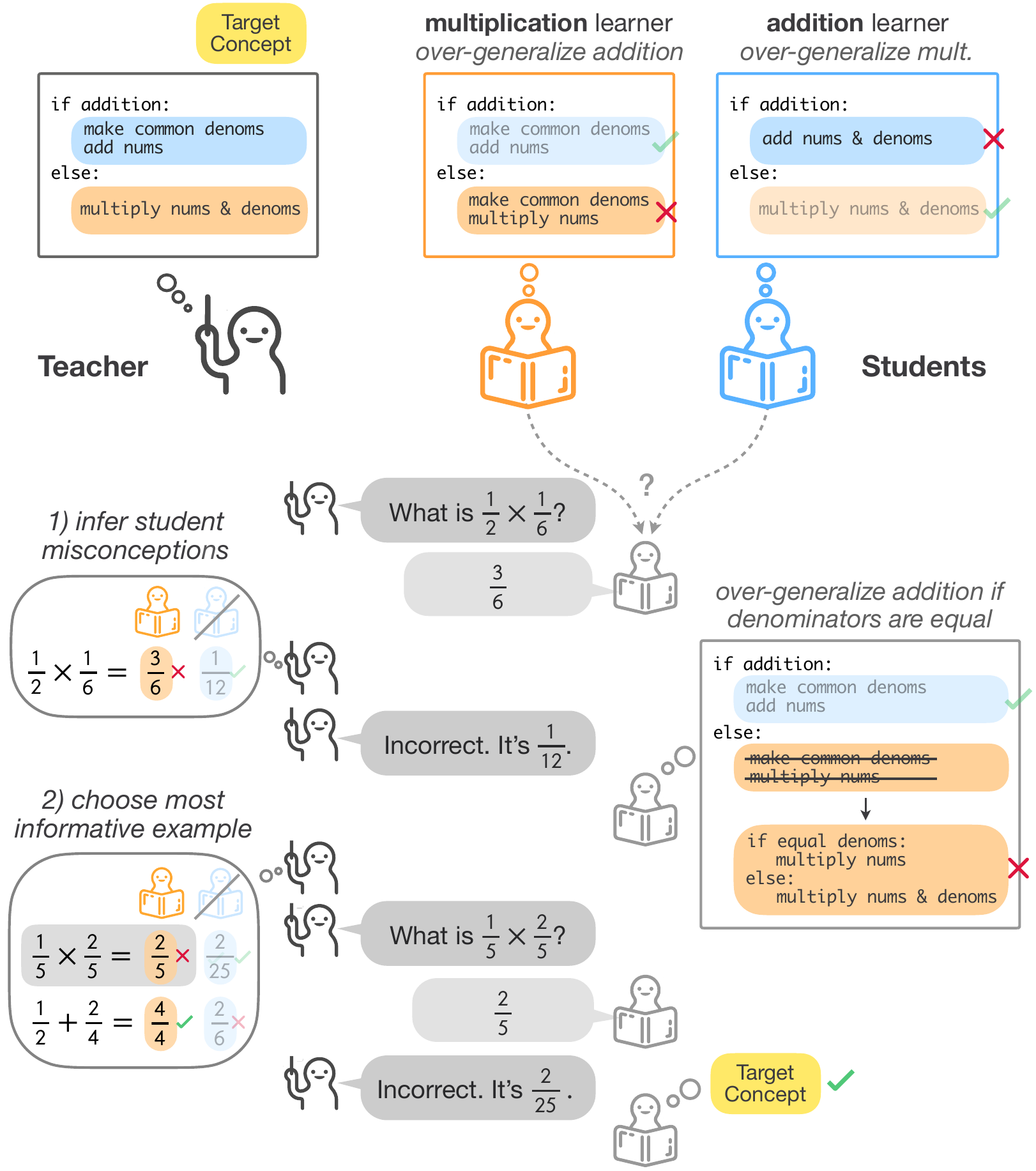} %
    \caption{
    In the \textbf{\benchmarkfull evaluation framework} \sectpar{s:benchmark}, a teacher selects examples to teach a target concept to a student; however, the student has prior misconceptions that are \emph{unknown} to the teacher. 
    In the fraction arithmetic task \sectpar{s:fraction_task}, some students (\emph{multiplication~learner}) tend to over-generalize the addition procedure of making common denominators and performing arithmetic only on numerators; others (\emph{addition~learner}) tend to over-generalize the multiplication procedure of applying arithmetic on both numerators and denominators. In order to teach effectively, the teacher must jointly 1) \emph{infer} the student's misconceptions online by observing their behavior throughout their interaction (\ie the teacher infers that the student is an \emph{addition~generalizer} after observing the prediction $\frac{1}{2} \times \frac{1}{6}\rightarrow\frac{3}{6}$), and 2) \emph{adapt} to such misconceptions by selecting examples that will most efficiently correct these misconceptions (\ie the teacher anticipates the student's new incorrect belief that all fractions {with equal denominators} should be treated as addition problems and selects the example $\frac{1}{5} \times \frac{2}{5}=\frac{2}{25}$ to correct it). 
    We propose \textbf{\adaptiveunknownteacher}, a two-part probabilistic approach that achieves adaptive teaching by maintaining explicit inferences about student priors \sectpar{s:adaptive_method}.
    }
    \label{fig:teaser}
\end{figure}

\begin{abstract}
When a teacher provides examples for a \mbox{student} to study, these examples must be \emph{informative}, enabling a student to progress from their current state toward a target concept or skill.
Good teachers must therefore simultaneously infer what students already know and adapt their teaching to students' changing state of knowledge. 
There is increasing interest in using computational models, particularly large language models, as pedagogical tools.
As students, language models in particular have shown a remarkable ability to adapt to new tasks given small numbers of examples. But how effectively can these models \emph{adapt as teachers} to students of different types?
To study this question, we introduce a suite of models and evaluation methods we call \benchmark. \benchmark has two components: (1) a collection of simulated Bayesian student models that can be used for  evaluation of automated teaching methods; (2) a platform for evaluation with human students, to characterize the real-world effectiveness of these methods. We additionally introduce
(3) \adaptiveunknownteacher, a new probabilistic model for adaptive teaching that jointly infers students' past beliefs and optimizes for the correctness of future beliefs. In evaluations of simulated students across three learning domains (fraction arithmetic, English morphology, function learning), \adaptiveunknownteacher systematically outperforms LLM-based and standard Bayesian teaching models. In human experiments, both \adaptiveunknownteacher and LLMs 
outperform non-adaptive random example selection.
Our results highlight both the difficulty of the adaptive teaching task and the potential of learned adaptive models for solving it.

\end{abstract}

\section{Introduction}
\label{s:intro}

Good teaching is \emph{adaptive} to students' specific beliefs and preconceptions \citep{sigma}. Imagine, for example, that you have been tasked with tutoring students fraction arithmetic. You may start by first probing a student's understanding,
asking them what $\frac{1}{5} \times \frac{2}{5}$ is. Suppose the student answers $\frac{2}{5}$. Immediately, you might develop a hypothesis about this student's misconceptions:
they seem to be over-generalizing the rule for \emph{addition}, only applying the operation to the numerator. Now suppose another student \emph{correctly} answers $\frac{1}{5} \times \frac{2}{5}=\frac{2}{25}$, but answers $\frac{1}{2}+\frac{2}{4}=\frac{2}{6}$. This student would appear to be over-generalizing the rule for \emph{multiplication}. These (discrete and systematic) categories of student misconceptions have been found to be widespread among real students learning fraction arithmetic \cite{fraction}.

As this example highlights, interactions with students reveal insights about their misconceptions, and these misconceptions in turn influence the course of effective teaching. 
A good teacher should provide different problems for these students to target their specific misconceptions: 
The addition-generalizer would benefit from multiplication examples, especially those with common denominators, while the multiplication-generalizer would benefit from addition examples.

What does this mean for NLP? Computational models---particularly language models (LMs)---are increasingly used as pedagogical tools
\citep{KASNECI2023102274}. But it is unclear how effectively any of today's models can tailor instruction
to perform
``in-context teaching'' \sectpar{s:problem} for students with differing degrees of skill and prior knowledge. In this work, we draw on a long line of literature on rational models of pedagogy \citep{Shafto2014} to study this question. 

To do so,
we introduce \textbf{\benchmarkfull}, an evaluation suite 
targeted at teaching students with varied prior misconceptions \sectpar{s:benchmark}. In \benchmark, a teacher is tasked with selecting examples to teach a particular target concept. As shown in Figure~\ref{fig:teaser}, the teacher selects examples one by one and can observe predictions made by the student. Importantly, the student has prior misconceptions that are unknown to the teacher. \benchmark is designed such that correct inferences about student misconceptions can enable more efficient learning.
\benchmark has two components:
\begin{enumerate}
    \item An \textbf{offline, probabilistic framework} for reproducibly evaluating how efficiently teachers can teach \textbf{simulated students} with unknown prior misconceptions \sectpar{s:synthetic_benchmark}.
    
    \item An evaluation platform for measuring teachers' efficacy with \textbf{human students} with prior misconceptions \sectpar{s:human_benchmark}.
\end{enumerate}

Simulated and human experiments allow us to characterize the pedagogical capabilities of teachers along several dimensions: 
their
inferences about student beliefs (\sect{s:gpt_student_inference}, \sect{ssec:human_student_accuracy}), adaptivity of chosen examples \sectpar{s:critical_examples}, and differences in teaching mathematical concepts \sectpar{ssec:gx_examples}. In addition to these evaluations, we introduce:
\begin{enumerate}
    \item[3.] A new probabilistic model, \textbf{\adaptiveunknownteacher (Adaptive Teaching tOwards Misconceptions)}, which performs \emph{online inference of student priors}, then uses these inferences to select informative teaching examples. \adaptiveunknownteacher provides proof-of-viability for adaptive teaching methods in simulated and human students \sectpar{s:adaptive_method}.
\end{enumerate}

Using \benchmark evaluations, we characterize \adaptiveunknownteacher, \gptfour, and a range of other methods. In simulated students, we find that while \gptfour exhibits some adaptation to student misconceptions, there is room for improvement with adaptive approaches;  it substantially underperforms \adaptiveunknownteacher, suggesting promise in using adaptive methods \sectpar{s:synthetic_experiments}. In human experiments, however, both \adaptiveunknownteacher and \gptfour outperform random example selection,
highlighting the potential of learned models (of various kinds) for adaptive teaching \sectpar{s:human_experiments}.\footnote{Our code is publicly available at \url{https://github.com/alexisjihyeross/adaptive_teaching}.}

\begin{figure*}[ht!]
    \centering
    \includegraphics[width=\textwidth]{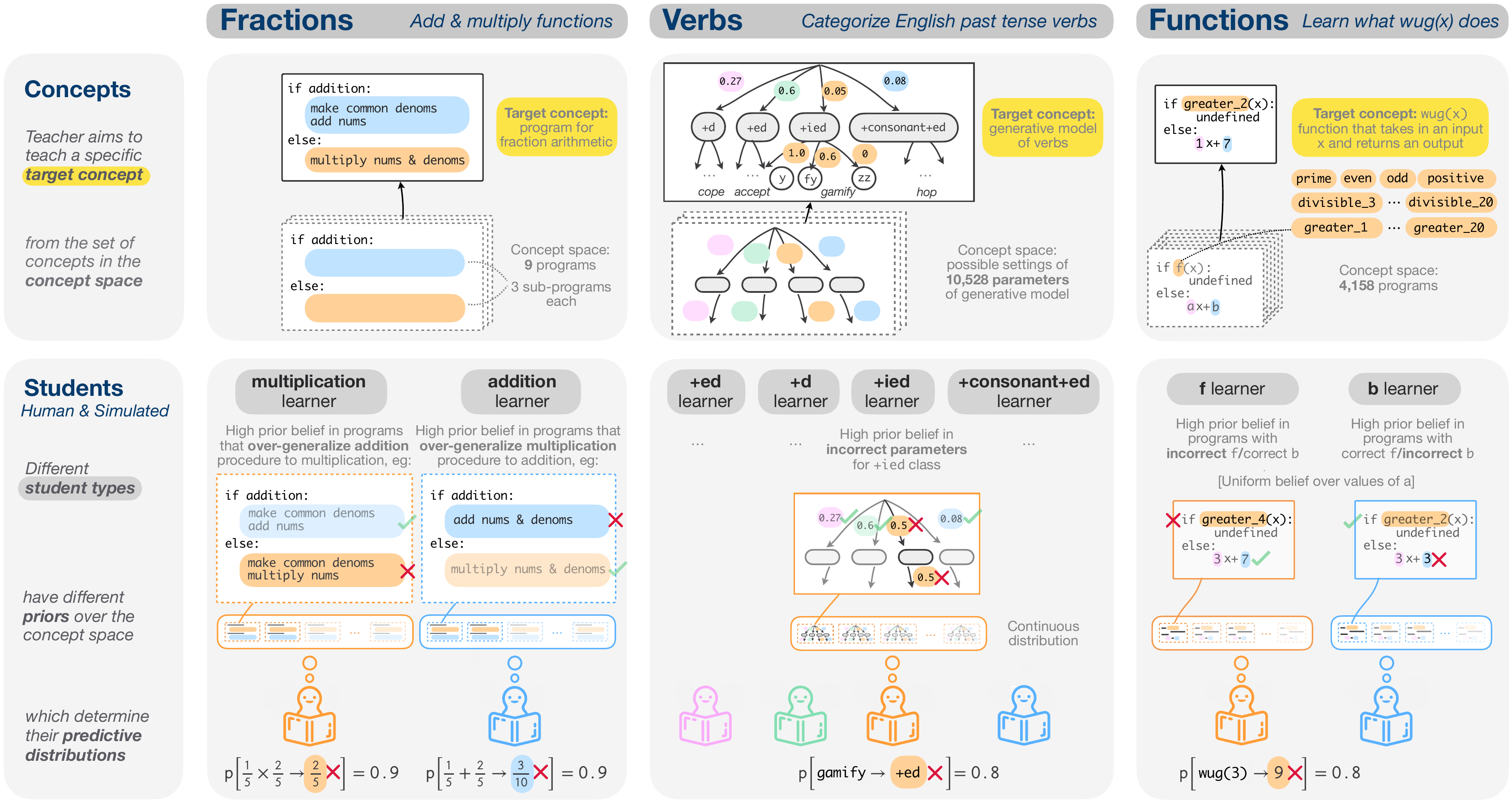}
    \caption{An overview of the tasks and student types in the \benchmarkfull evaluation framework \sectpar{s:benchmark}. \benchmark has three tasks: fractions, verbs, and functions. For the fraction and function tasks, a student's concept space consists of \emph{programs}; for the verbs task, a student's concept space is the space of \emph{generative models} corresponding to English past tense verb classes. 
    }
    \label{fig:benchmark}
\end{figure*}

\section{Preliminaries}
\label{s:problem}
\label{s:prelims}

We formulate our problem setting as one in which a \textbf{teacher} interactively provides feedback to a \textbf{student} to communicate a new concept (a mapping from inputs $x$ to outputs $y$). A teacher observes a sequence of $(x, \hat{y})$ pairs guessed by a student, and must infer what additional training example $(x, y)$ pair will most improve the student's 
understanding of the concept.
We term this problem \textbf{in-context teaching} to draw an analogy to the widely studied phenomenon of ``in-context learning'' \citep{icl,min-etal-2022-rethinking,what-learning-icl}. Formally:
\begin{itemize}
    \item The teacher begins with a target \textbf{concept} $\conceptmath^*$ drawn from some concept space $\fullconceptmath$. A concept parameterizes a mapping between an input space $\Xspace$ and an output space $\Yspace$.
    In Fig.~\ref{fig:teaser},
    $\conceptmath$ is the true procedure for adding and multiplying fractions, and $\fullconceptmath$ is the space of other possible fraction manipulation algorithms, so $\Xspace$ contains arithmetic expressions involving fractions, and $\Yspace$ contains fractions.

    \item We assume that teachers interact with students by providing \textbf{examples} $(x_i, y_i) \in \Xspace \times \Yspace$. For convenience, we denote a sequence of such examples $(\X, \Y) = [(x_1, y_1), \ldots, (x_n, y_n)]$. In each round of teaching, the teacher first presents the student with an input $x_i$, the student produces a guess $\hat{y}_i$, and then the teacher reveals the true $y_i$.\footnote{This is both a model of real-world educational practice and a standard paradigm for online learning; future work might study richer forms of interaction with explanations and instructions.}

    \item Given a collection of examples $(\X, \Y)$, we assume that students compute a \emph{posterior} over concepts $p_S(\conceptmath \mid \X, \Y)$. 
    For example, a student who just seen that $\frac{1}{3} \times \frac{2}{3} = \frac{2}{9}$ may be less likely to believe that fraction addition and multiplication follow the same rules.
    The process by which students infer concepts from examples can in principle be arbitrary; however, for some models in this paper we will assume that students are Bayesian, with: \begin{equation}\label{eq:bayesian-student} p_S(\conceptmath \mid \X, \Y) \propto p_S(\conceptmath) \prod_i p_S(y_i \mid x_i, \conceptmath) \end{equation} under some \textbf{prior} belief $p_S(\conceptmath)$ 
    about the concept space.
\end{itemize}
Given this setup, a teaching strategy is a \textbf{policy} $p_T(x, y \mid \X, \Y)$ that chooses examples to maximize the probability that the student assigns to $\conceptmath^*$. In the \textbf{optimal teaching (OT)} framework of \citet{Shafto2014}, for example, $p_S(\conceptmath \mid \X, \Y)$ is assumed known, and teachers choose examples:
\begin{equation}
\label{eq:example_selection}
    x_{i+1}, y_{i+1} = \argmax_{x, y} ~ p(\conceptmath^* \mid \X, \Y, x, y)
\end{equation}
In the running example, this criterion is more likely to select examples of addition for a student who has already mastered multiplication, and vice-versa.
In addition to this greedy approach, it is possible to \emph{plan} sequences of informative examples for students \citep{Rafferty2016-tp}.

As discussed in \sect{s:intro}, however, the assumption that teachers have exact knowledge of $p_S(\conceptmath \mid \X, \Y)$ is often unrealistic---real-world teaching involves students of many different types, whose beliefs and misconceptions may not be known (or easily discerned) \emph{a priori}. Thus, we study teaching when students' priors are themselves unknown. We assume that students are drawn from a distribution over \textbf{student types} $p(\alpha)$, each associated with a concept prior $p(\conceptmath \mid \alpha)$. In the running example, these priors may correspond to different beliefs about the algorithms for fraction addition and multiplication, with ``addition generalizers'' assigning high probability to a spurious multiplication algorithm that manipulates only numerators.

In this setting, teachers must still implement an effective example selection policy $p_T(x, y \mid \X, \Y)$; however, choosing informative examples now requires \emph{inferring} student priors in order to estimate the effect of these examples on $p_S(\conceptmath^* \mid \X, \Y)$. In the next section, we describe our proposed framework for evaluating adaptive teaching policies. In \sect{s:models}, we describe a set of candidate adaptive teaching policies (including our new \adaptiveunknownteacher model), and in \sect{s:synthetic_experiments} and \sect{s:human_experiments} use \benchmark to evaluate these teaching policies with simulated and human students.

\section{The \benchmark Evaluation Framework}
\label{s:benchmark}

\benchmark has two parts: an offline evaluation framework with simulated students \sectpar{s:synthetic_benchmark}, and a platform for doing experiments with human students \sectpar{s:human_benchmark}. We first describe the tasks in \benchmark \sectpar{s:tasks}. An overview of \benchmark is shown in Figure~\ref{fig:benchmark}.

\subsection{Tasks}
\label{s:tasks}

\paragraph{Fractions}
\label{s:fraction_task}

In the first task, the teacher aims to teach the student how to add and multiply fractions. Here, student types correspond to different prior beliefs (possibly incorrect) about the rules for fraction arithmetic.

\paragraph{Verbs}
\label{s:verb_task}

The second task is English past-tense conjugation. 
In this task, students are presented with lemmas and must choose an appropriate ending (\eg \emph{play} $\to$ \emph{+ed}, fry $\to$ \emph{+ied}). Here, student types correspond to different degrees of familiarity with possible English past-tense endings.

\paragraph{Functions}
\label{s:function_task}
In the third task, reminiscent of existing number concept learning tasks \citep{number_game}, 
the teacher aims to teach the student 
a function that takes in numbers and returns either numbers or \undefined. These functions can be represented as \emph{programs} that take an input \functioninput and compute:
\begin{align}
  \texttt{if f(x)}&\texttt{: return undefined} \label{eq:function_prog_form} \\
  \texttt{else}&\texttt{: return a*x+b} \nonumber
\end{align}
where \fnox is a boolean function and \ga, \gb are integers. The teacher chooses input/output pairs (\functioninput, \wugx) to show the student to maximize the student's belief that the concept is the correct program \goldconcept. We create 24 target concepts, which combine 3 unique settings of \ga/\gb and 8 settings of \fnox. 
Student types can be instantiated by selecting {preferences for specific primitives} (\eg \fnox, \ga, \gb); arbitrary priors over programs can then be derived from these preferences for primitives.

\subsection{Simulated Students}
\label{s:synthetic_benchmark}

The first component of \benchmark evaluates teachers with simulated, Bayesian students. These students maintain belief distributions over the full concept space.\footnote{The concept space consists of 9 concepts for the fraction task, 4,158 concepts for the function task, and a continuous space of possible values for 10,528 parameters (corresponding to 329 features and 4 verb classes) for the verb task.} 
As shown in Figure~\ref{fig:benchmark}, different ``student types'' are implemented by initializing students with different priors over the concept space. 
All student types begin with low initial belief in the target concept \goldconcept and assign high probability to other spurious concepts.

\paragraph{Fractions}
For the fraction task, we represent understanding of fraction arithmetic as programs. Students maintain a belief distribution over the space of possible programs, as shown in Figure~\ref{fig:benchmark}. We create two student types, \textbf{\addgeneralizer} (a model of a student who has not yet mastered multiplication and incorrectly applies the procedure for addition to multiplication) and \textbf{\multgeneralizer} (a student who performs addition by incorrectly applying the procedure for multiplication). These correspond to common incorrect strategies that children exhibit when learning fraction arithmetic \cite{fraction} by \emph{over-generalizing} the procedure for operation to another.

\paragraph{Functions}
For the function task, students again maintain a belief distribution over the space of possible concepts.
We create two types of students for each target concept: a {\fknower} and a {\gknower}. The \textbf{\gknower} knows the true value of \gb but has an incorrect, spurious belief about what function \fx is; the \textbf{\fknower} knows the true \fx in the target program \goldconcept but has an incorrect belief about the value of \gb. 
See \sect{s:incorrect_concepts} for how we select incorrect beliefs for students.

\paragraph{Verbs}
For the verb task, we represent understanding of verb conjugation as \emph{generative models} of English past-tense verbs. 
Students are 
na\"ive Bayes models with features for word-final character n-grams, so $p(\conceptmath \mid \X, \Y)$ is a distribution over model parameters, with Dirichlet/Beta priors over the verb class/feature occurrence parameters, respectively. 
We fit
a na\"ive Bayes model on the 
Unimorph dataset\footnote{\url{https://github.com/unimorph/unimorph}} \citep{batsuren-etal-2022-unimorph}
and use the mode of the resulting posterior as the target concept.\footnote{By Dirichlet--Multinomial conjugacy, the posterior distribution over parameters factorizes, is also a product of Dirichlet and Beta distributions, and can be efficiently computed.}

We create four student types by picking one of the classes as the ``unknown'' class: a \textbf{\dnotknower}, \textbf{\ednotknower}, \textbf{\consonantednotknower}, and a \textbf{\ytoiednotknower}. To simulate students who are familiar with all but one class, we initialize the student's priors by using the posterior mode parameters of the model fit on the full data, setting the parameters for the ``unknown'' class to all 1s (effectively removing any learned information about the class): Figure~\ref{fig:benchmark} shows how setting the prior in this way determines the mode of the \ytoiednotknower's prior distribution over generative models.

\subsection{Human Students}
\label{s:human_benchmark}

The second component of the \benchmark evaluation framework is a platform for evaluating adaptive teaching with human students, specifically for the function learning task. Human participants are tasked with learning what a ``mystery machine'' called \wug does. They are given 10 minutes to interact with a teacher who presents teaching examples through a chat interface. Their task is to figure out when \wugx is undefined (\ie guess what \fnox is), and when \wugx is defined, what it computes (\ie what \ga and \gb are in \linearconcept). They can submit guesses for how \wugx operates whenever and however many times they choose to during the 10 minutes of interaction. 
\wug guesses have 3 components corresponding to \fnox, \ga, and \gb, and we allow partial guesses. See \sect{s:human_setup} for more details on instructions, bonus compensation, and other parts of the human study.

We create \fknowers and \gknowers by priming the human participants with hints from a ``Dr. Smith''; \fknowers receive a hint with the correct \fnox but incorrect value for \gb, and the reverse for \gknowers. An example hint is given in Table~\ref{tab:human_hints}.

\begin{figure*}[ht!]
    \centering
    \includegraphics[width=0.85\textwidth]{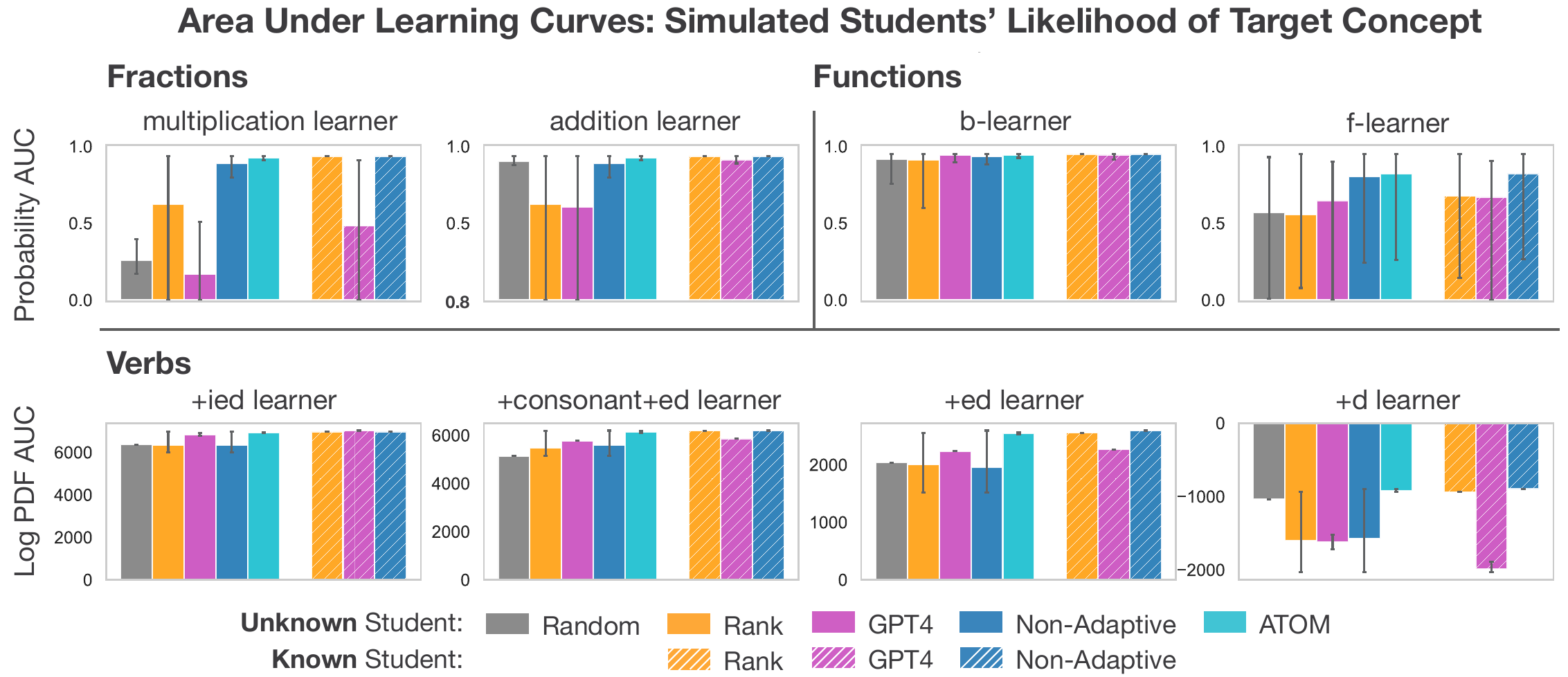}

    \caption{Area under simulated learners' learning curves, where curves plot learners' posterior beliefs in the target concept by number of datapoints. We report results by task and student type with 3 random seeds per bar. Dashed bars indicate that the true student type is assumed. Note that the y-axis for the \gknower for functions starts at 0.8, as these learners all learn the concept early on, and so differences in teaching methods are small. 
    Error bars show min/max values across seeds.
    Full learning curves are shown in Figure~\ref{fig:synthetic_learning_all_curves}.}
    \label{fig:synthetic_learning_all_bar}
\end{figure*}

\begin{figure*}[h!]
    \centering
    \includegraphics[width=\textwidth]{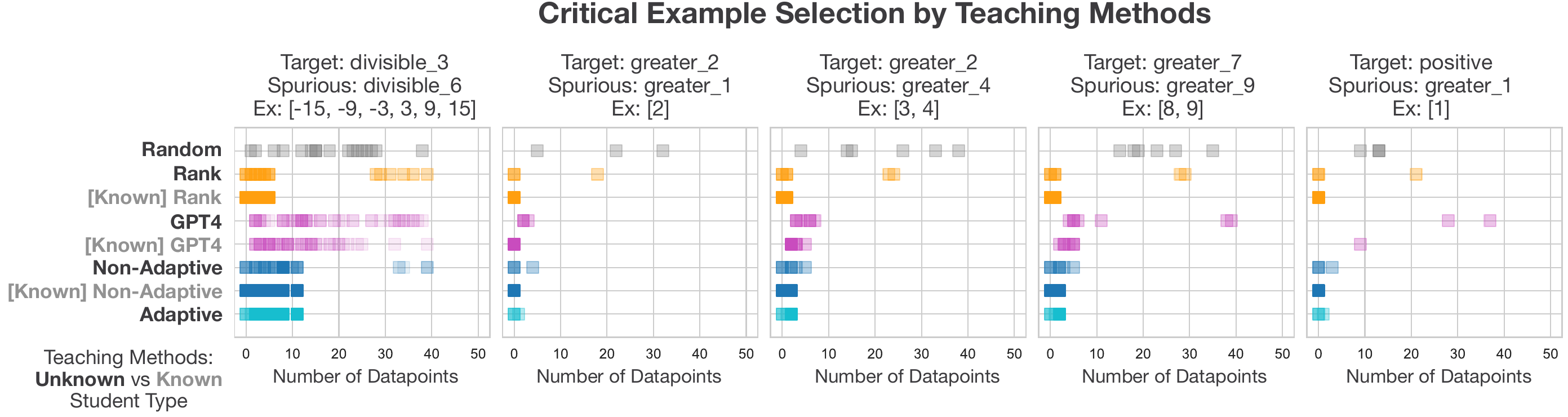}
    \caption{Critical example selection by different teaching methods for the function task. Results are for simulated \gknowers, who have a spurious belief about \fnox that agrees with the target \fnox on all but a few examples, as labeled. The opacity of each square corresponds to the mean value of whether the example chosen by the teaching method at that step in learning is a ``critical example'' (averaged across experimental conditions: seed and concepts). See \sect{s:critical_examples} for details. We report a subset of results here; see \sect{fig:critical_examples_full} for full results.
    }
    \label{fig:critical_examples}
\end{figure*}

\section{Models}
\label{s:methods}
\label{s:models}

\subsection{Teaching Toward Misconceptions}
\label{s:adaptive_method}
We introduce an approach that makes explicit inferences about the parameters of the student's prior. We call this method \textbf{\adaptiveunknownteacher (Adaptive Teaching tOward Misconceptions)}.

Like the OT method described in \sectpar{s:prelims}, \adaptiveunknownteacher assumes that students are Bayesian reasoners and chooses examples to maximize the posterior probability that the student assigns to the target concept $\conceptmath^*$. Because the student's prior is unknown, however, this process involves two steps:

\begin{enumerate}
    \item \textbf{Maximum \emph{a posteriori} estimation of student priors}. Recall that, during interaction, the teacher provides inputs $x_i$, then observes student guesses $\hat{y}_i$ before providing ground-truth labels $y_i$. In \adaptiveunknownteacher, the teacher selects student prior parameters that best explain the student's sequence of guesses:
    \begin{align}
    \hspace{-1em}
        \alpha_i &= \argmax_\alpha \sum_i \log p_S(\hat{y}_i \mid \X, \Y, x_i, \alpha)
    \end{align}
    where $(\X, \Y) = [(x_1, y_1), \ldots, (x_{i-1}, y_{i-1})]$
    Estimating this $\argmax$ requires a tractable procedure for computing the posterior predictive distribution, which is available for each simulated student model we evaluate in \sect{s:synthetic_experiments}.

    \item \textbf{Optimal selection of informative examples}. As in OT, once $\alpha$ has been estimated, we choose an example $(x_{i+1}, y_{i+1})$ to optimize:
    \begin{align}
    \hspace{-1.3em}
        \argmax_{x_{i+1}, y_{i+1}} p_S(\conceptmath^* \mid \X, \Y, x_i, y_i, x_{i+1}, y_{i+1}, \alpha)
    \end{align}
\end{enumerate}

We note that many more sophisticated \adaptiveunknownteacher-type methods are possible---for example, explicit marginalization (rather than MAP estimation) of student priors. More basically, the method described above does not perform any active experimentation to identify the student prior; alternative \adaptiveunknownteacher implementations could explicitly trade off between exploration (of the student type) and exploitation (of the student posterior).

\subsection{Other Models}
\label{s:baselines}

\paragraph{Random} The \textbf{\randomteacher} baseline uniformly samples an input to show the student. 

\paragraph{Ranking} A second baseline ranks the datapoints at the first step according to the objective in Eq~\ref{eq:example_selection}, then chooses them in this order for the rest of the teaching interaction. The student type is chosen uniformly. We refer to this baseline as \textbf{\rankingunknownteacher}.

\paragraph{Non-Adaptive} A third baseline selects examples according to the OT objective in Eq~\ref{eq:example_selection} but maintains a \emph{fixed} guess about the student type, chosen uniformly at the start of teaching. This baseline can be thought of as an ablation of the adaptive piece of \adaptiveunknownteacher. We refer to this baseline as \textbf{\greedyunknownteacher} \citep{Shafto2014}.

\paragraph{GPT-4} We prompt the \texttt{gpt-4-0314} model to select teaching examples (and provide no other explanations); the prompt describes the target concept, the student's hypothesis space, and the student types. The model is instructed to try to infer the student type in order to teach most efficiently. See Appendix~\ref{s:prompts} for actual prompts.
To control for the fact \gptfour sometimes generates incorrect outputs for examples, we use \textbf{ground truth} outputs for generated inputs.\footnote{We parse \gptfour generations to get inputs and create new messages with target outputs. More details can be found in~\sect{s:prompts}.} We call this method \textbf{\gptfourunknownteacher}.

\subsection{Oracle Models}

We also compare against several methods that assume access to the true student. These serve as comparison points for how well methods \emph{could} do if they inferred the correct student model. We run this reference for all methods except \randomteacher, which does not make use of a student model. We refer to these methods as \textbf{\rankingknownteacher}, \textbf{\greedyknownteacher}, and \textbf{\gptfourknownteacher}.

\section{Simulated Experiments}
\label{s:synthetic_experiments}

\subsection{Experimental Set-Up}
\label{s:experiment_set_up}

We run 3 random seeds for all experimental conditions. For all methods except the \gptfour-based methods, we restrict the teaching methods from selecting previously selected examples. Teaching interactions last 40 steps for the fraction/function tasks and 50 steps for the verb task.
For non \gptfour methods, we enumerate over either the full dataset (fractions/functions) or a sampled subset (500 examples for verbs) to choose teaching examples.

\subsection{Students' Learning Efficiency}
\label{s:results_learning_efficiency}

We evaluate teacher effectiveness by measuring 
the student's \textbf{probability of the target concept, \goldconcept}. 
Figure~\ref{fig:synthetic_learning_all_bar} shows the area under simulated students' learning curves, where curves reflect students' beliefs in \goldconcept (full curves are shown in Figure~\ref{fig:synthetic_learning_all_curves}). We observe that \adaptiveunknownteacher performs almost as well as the optimal strategy, \greedyknownteacher, and outperforms \greedyunknownteacher, suggesting adaptation is both possible (\ie student type is inferrable from interaction) and that it leads to improved teaching efficiency.
We also observe that both \gptfourknownteacher and \gptfourunknownteacher outperform 
\randomteacher
but underperform both ATOM and the non-adaptive probabilistic approaches. 

\subsection{\gptfour's Inferences about Student Type}
\label{s:gpt_student_inference}

We query \gptfourunknownteacher for the student type at the end of the teaching interaction (\emph{Based on this interaction, which kind of student do you think I was at the start of this teaching session \dots}). See Table~\ref{tab:function_prompts_student} for an example prompt. The mean accuracies of \gptfour's student type guesses 
are \textbf{100\%} for verbs, \textbf{66.67\%} for fractions, and \textbf{53.47\%} for functions.

A possible explanation for these discrepancies is that for the fraction and function tasks, students successfully learn the target concept 
by the end of the teaching interaction and thus make accurate predictions; for the verb task, however, students are still making errors by the end.\footnote{See Figure~\ref{fig:student_acc_teaching_examples} for how the correctness of student predictions on teaching examples changes throughout learning.}
We analyze how these accuracies change throughout the teaching interactions.
For the function task, the student type accuracies are \textbf{64.2\%}, \textbf{60.4\%}, \textbf{56.3\%}, \textbf{53.47\%} after 10, 20, 30, and 40 steps, respectively: This decrease suggests that \gptfour exhibits \textbf{recency bias} in making inferences about student type.

\subsection{Selection of Critical Examples}
\label{s:critical_examples}

For function concepts, we evaluate how early teaching methods select ``critical examples,'' or key examples that distinguish the target \fnox from the spurious \fnox.\footnote{We consider a subset of target concepts/spurious concepts where the number of critical examples is less than 10.} Consider the case where the target \fnox is \f{greater\_2} but the \gknower believes it is \f{greater\_4}: The critical examples are \textbf{3} and \textbf{4} because they are the only examples on which the target \fnox and spurious \fnox would return different outputs. 
Observing \wugx on one of these inputs would make clear to the \gknower that their belief about \fnox is wrong. Therefore, an effective teacher should select such examples early in teaching.

As shown in Figure~\ref{fig:critical_examples}, the probabilistic methods 
assuming
the student type (\rankingknownteacher, \greedyknownteacher) all select critical examples early. \gptfourknownteacher also shows a concentration of critical examples early, though 
they
are more spread out for some concepts (\ie for target \fnox \f{divisible\_3}/\f{positive}); \gptfour thus exhibits some pedagogical reasoning, focusing on examples that will target the \gknower's misconceptions when it knows that the student type is a \gknower.
When \gptfourunknownteacher does not know the student type, we observe that critical examples are still more concentrated at the start than for \randomteacher, suggesting some degree of adaptivity.
Finally, we observe that \adaptiveunknownteacher selects critical examples at comparable points to \greedyknownteacher despite having to guess about student type.

\begin{figure}[b!]
    \centering
    \includegraphics[width=\columnwidth]{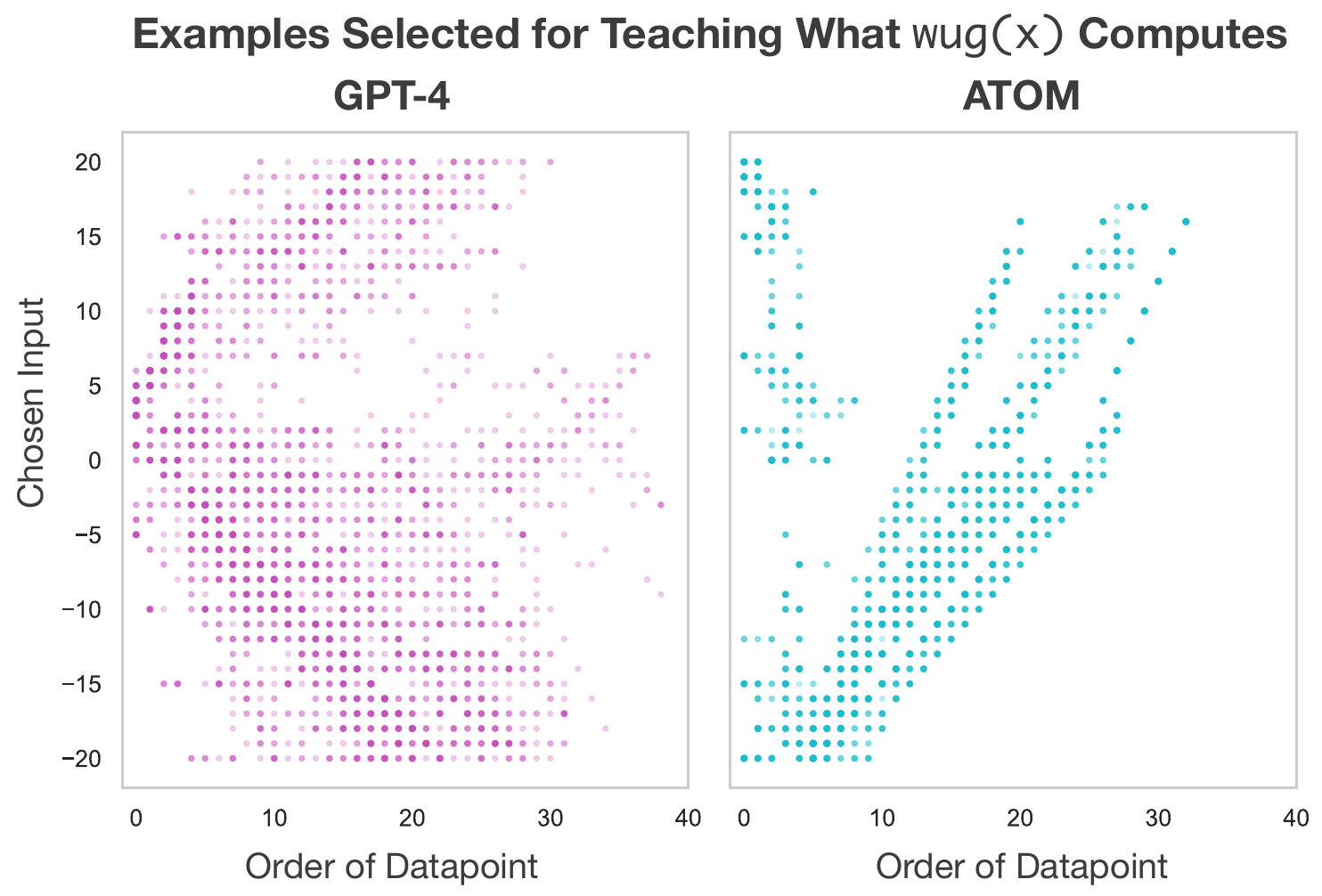}
    \caption{Examples selected by different teaching methods for teaching \ga and \gb in the function learning task (\ie what \wugx computes when it is defined). 
    The x-axis indicates the order of the chosen example compared to other examples targeting \ga and \gb. 
    }
    \label{fig:gx_examples_synthetic}
\end{figure}

\subsection{Qualitative Differences in Teaching Math} 
\label{ssec:gx_examples}

For function concepts, recall that when \wugx is defined, it computes \texttt{a*x+b}. We analyze 
how different methods teach
what \ga and \gb are when \wugx is defined by plotting the inputs they choose.
As Figure~\ref{fig:gx_examples_synthetic} shows, \gptfourunknownteacher tends to select inputs in order of increasing magnitude. In contrast, \adaptiveunknownteacher starts with higher-magnitude examples, then selects examples in increasing order. These qualitative differences suggest that \gptfour may have encoded information that inputs closer to the origin are easier to learn from than those further from the origin.\footnote{In contrast, \adaptiveunknownteacher scores inputs according to how many incorrect concepts they ``rule out,'' treating all else as equal. 
Because higher-magnitude inputs tend to result in outputs that can be explained by fewer functions, \adaptiveunknownteacher selects higher-magnitude inputs early in teaching.
}

\begin{figure}[t!]
    \includegraphics[width=0.75\columnwidth]{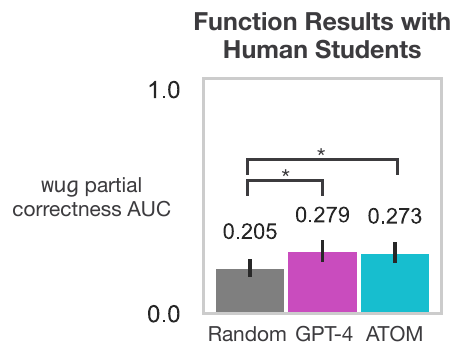}
    \caption{Results with human students showing how efficiently students guessed the correct \wug concept \sectpar{s:human_learning_efficiency}. Stars indicate statistically significant results under a paired t-test. Error bars show 95\% confidence intervals.}
    \label{fig:human_bar}
\end{figure}

\section{Human Experiments}
\label{s:human_experiments}

\subsection{Experimental Set-Up}
\label{s:human_setup}

We recruit Prolific users who are fluent English speakers and who indicated some experience with computer programming. We pay participants a base pay of $\$4.00$ per study ($\$16$/hour) and offer a bonus based on the accuracy of their predictions to the teacher and on how early they guess the correct value for \wugx. See Appendix~\ref{s:human_appendix} for details.

We evaluate \randomteacher, \gptfourunknownteacher, and \adaptiveunknownteacher teachers on the function task. We run 5 experiments per teaching method per experimental condition.\footnote{There are 22 unique experimental conditions (11 target concepts, 2 student types per concept).} These experiments were classified as an exempt Benign Behavioral Intervention by our IRB.

\subsection{Students' Learning Efficiency}
\label{s:human_learning_efficiency}
We evaluate the effectiveness of teachers by measuring the \textbf{correctness of \wug guesses} made by human participants, computing an AUC-like metric. We consider all timestamps where at least one participant made a guess for \wug and compute a partial correctness metric for each guess: $(\text{f is correct}) + 0.5 (\text{a is correct}) + 0.5 (\text{b is correct})$.\footnote{If no new guess was made by a user, we use their last guess.} We report the \emph{mean} correctness values across timestamps.

As shown in Figure~\ref{fig:human_bar}, we find that both \gptfourunknownteacher and \adaptiveunknownteacher improve significantly over the \randomteacher baseline
($p < 0.05$ using a paired t-test). 
Interestingly, these differences are entirely explained by model behavior for \gknowers: after controlling for student type, improvements over the random teacher are significant for \gknowers but 
insignificant
for \fknowers. For results by student type, time, and individual participants, see \sect{s:human_appendix_results}.

\subsection{Inferences about Student Type}
\label{ssec:human_student_accuracy}

We evaluate the accuracy of predictions of student type made by \adaptiveunknownteacher and \gptfourunknownteacher after each minute of teaching. 
\adaptiveunknownteacher makes more accurate predictions than \gptfourunknownteacher, with respective mean accuracies of \textbf{71.33\%} and \textbf{52.61\%} across participants and minutes. Accuracies over time are shown in \sect{s:human_appendix_results} (Figure~\ref{fig:human_student_accuracy}).

\section{Discussion}
\label{s:discussion}

Both our human and simulated results show that \adaptiveunknownteacher and \gptfourunknownteacher exhibit pedagogical ability over random example selection.
\adaptiveunknownteacher's performance with human students suggests that 
the Bayesian assumptions made by \adaptiveunknownteacher are accurate models of some aspects of human learning. Across human and simulated experiments, we find evidence of some adaptivity in \gptfour, though less than in \adaptiveunknownteacher, both in the examples selected \sectpar{s:critical_examples} and inferences about student type (\sect{s:gpt_student_inference}, \sect{ssec:human_student_accuracy}). See \sect{s:results_adaptation} for additional analyses about how adaptive selected teaching examples are to student beliefs.

We also observe other qualitative limitations of \gptfour as a teacher: selecting the same teaching examples multiple times or terminating teaching early due to an incorrect belief that all teaching examples have been exhausted. It is important to highlight that because we use ground truth outputs (\eg ground truth function evaluations) with \gptfour, the \gptfour results represent an upper bound on 
\gptfour's performance.

Despite these limitations, however, \gptfour performs comparably to \adaptiveunknownteacher with human students, suggesting pedagogical benefits beyond adaptivity. In particular, the analysis in \sect{ssec:gx_examples} suggests that language models may encode information about human learning that is hard to represent in more structured models like \adaptiveunknownteacher---\eg that it is easier for humans to learn the weights of a line from inputs closer to the origin.

Together, our results point to complementary advantages of LM teachers like \gptfour and more structured, probabilistic methods like \adaptiveunknownteacher. They suggest that there is substantial headroom to improve real-world teaching by augmenting the inferences of structured models with richer information about the priors that humans bring to learning, whether by combining such structured methods with information encoded in LMs or by developing other rich models of student priors---\eg by learning more complex ``student types'' from naturally occurring data. We perform an initial experiment with human students
in this direction 
by combining \gptfour and \adaptiveunknownteacher but do not find improvements compared to either teacher alone; see \sect{s:gpt_atom_combination} for details.

Other directions for future work include modeling more complex student phenomena---accounting for students who ask questions, reason pedagogically about teacher intentions, and provide feedback to teachers \citep{Chen2022-jw}---and creating methods for adaptive teaching with natural language explanations
in real-world teaching domains. 

\section{Related Work}

This work builds on a long line of work in \textbf{rational models of pedagogy}, including Bayesian models like those described by \citet{Shafto2014} and \citet{Sumers2021-rt}, as well as improved planning and inference procedures like the one described by \citet{Rafferty2016-tp}. Past work generally assumes students' initial belief states are known.
In parallel, \citet{Rafferty2015-dc} use an inverse planning model to infer students' prior beliefs from their actions, and \citet{Chen2022-jw} show that human teachers adapt examples to these prior beliefs.

This work is also closely related to other bodies of work that aim to infer 
student knowledge.
\textbf{Item response theory (IRT)} infers a scalar measure of student's skill based on their responses to questions \citep{lord, hambelton2003irt}.
\textbf{Knowledge tracing (KT)} models students' evolving knowledge states over time separately for a fixed set of skills \citep{knowledge-tracing}; previous work has used both bayesian methods for modeling individual students' prior knowledge \citep{bayesian-knowledge-tracing} and neural models for modeling \citep{deep-knowledge-tracing} and targeting \citep{srivastava-goodman-2021-question} students' evolving learning states. 
In contrast to IRT and KT, our work aims to infer the learner’s entire prior and posterior over the concept space; these inferences in turn enable more fine-grained design of individual teaching examples. 

Inferring student misconceptions from errors also uses tools from a broader literature on computational models of \textbf{theory of mind}. Prominent work 
includes general-purpose bayesian models of other agents' beliefs and desires \citep{Baker2011-yv} and models of pragmatic inference grounded in recursive reasoning about speakers and listeners \citep{pragmatic_games}. More recent work has studied theory of mind capabilities in LMs;
they find largely negative results in unaugmented LLMs \citep{sap2023neural} but positive results from LMs augmented with structured belief representations \citep{sclar2023minding}. Recent work has also explored LLMs' theory of mind abilities in teaching smaller LMs \citep{saha2023language}.

There is also a large body of work on how to optimally provide and interpret human \textbf{supervision for ML models}. General frameworks for this problem include Machine Teaching \citep{machine_teaching} and Cooperative IRL \citep{Hadfield-Menell2016-jw}; related ideas appear in program synthesis \citep{Vaduguru2022-ni}, robot learning \citep{Milli2019-mm, 10.5555/2447556.2447672}, and natural language processing \citep{Li2023-ds} as well.

There has been increased interest in using LLMs to assist or supplement human teachers. See \citet{KASNECI2023102274} for a survey and \citet{wang2023stepbystep} for a specific application to math teaching problems.
Concurrent work by \citet{chandra2024wat} uses program synthesis techniques to infer misconceptions and provide explanations about Javascript. Our work adds to this literature by providing a framework that allows reproducible evaluation of the effectiveness and personalization skills of LLMs as teachers, as well as a new model that empirically improves upon LLM baselines in teaching humans a new task.

\label{s:related}

\section{Conclusion}
\label{s:conclusion}

We introduce \benchmark, an evaluation suite measuring how effectively automated teaching methods can teach students with different prior misconceptions. We also introduce \adaptiveunknownteacher, a two-part probabilistic approach to adaptive teaching that maintains explicit inferences about student priors. Our evaluations of \adaptiveunknownteacher, LLMs, and other probabilistic baselines  with both simulated and human students highlight the potential of learned adaptive models for solving the adaptive teaching task.

\section{Limitations}

One limitation of our work is that in our model of teaching, teachers are limited to example selection and students are limited to observation. These restrictions leave out that teachers can provide explanations, and students can ask questions and provide feedback to teachers. An interesting direction for future work would be to both create an evaluation framework for such phenomena and develop models for these richer forms of teaching and learning feedback. Relatedly, students can also engage in pedagogical reasoning about why teachers chose particular examples, which can in turn influence how they learn from these examples; while the simulated students in \benchmark do not model this phenomenon, future work could explore richer models of students.

Another limitation is that we create the student types by instantiating priors in particular ways rather than deriving the priors from real-world data. Future work could explore how to automatically \emph{learn} the types of priors that human students bring to different teaching scenarios.

\section*{Acknowledgements}
We are grateful to Ekin Akyürek, Leshem Choshen, Gabriel Grand, Robert Hawkins, Andi Peng, Megha Srivastava, Lionel Wong, Zhaofeng Wu, and members of the MIT Language \& Intelligence group for helpful discussions and feedback on this work. AR was funded by the NSF GRFP 2023357727; additional funding was provided by the National Science Foundation under grants IIS-2238240 and IIS-2212310, and by the Alfred P.\ Sloan Foundation.

\bibliography{custom}

\appendix

\section{Do teaching examples target student misconceptions?}
\label{s:results_adaptation}

\begin{figure*}[t!]
    \centering
    \includegraphics[width=\textwidth]{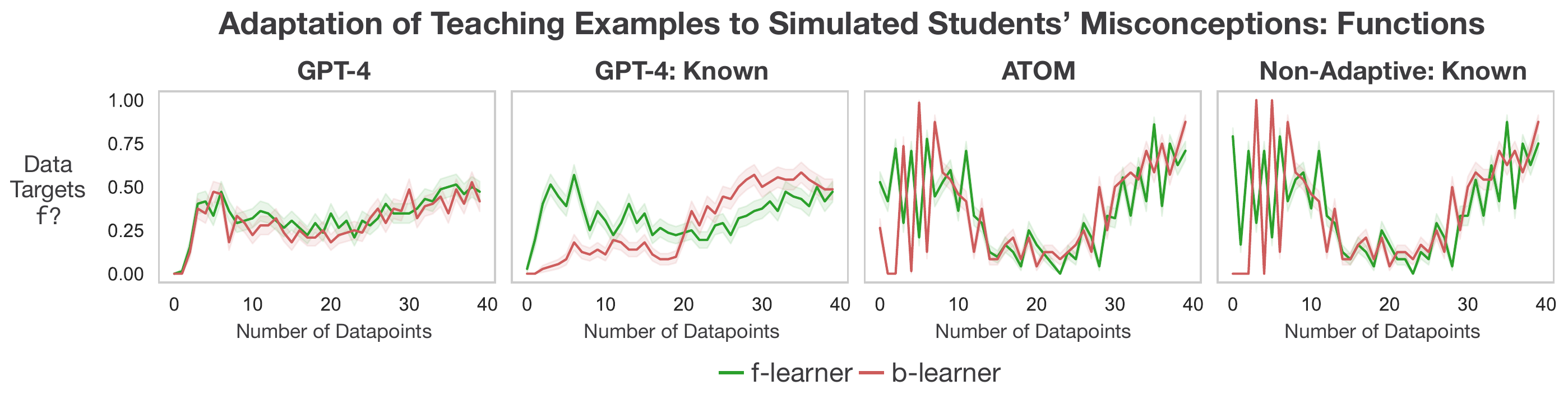}
    \caption{Plot showing whether the data chosen by teaching methods target learning \fnox (\ie when \wugx is undefined) or \linearconcept (\ie what \wugx computes when defined) for the function task. An input x for which \wugx is \emph{undefined} targets \fnox, and an input for which \wugx is \emph{defined} targets \linearconcept. Intuitively, \gknowers benefit more from seeing examples targeting \fnox early on in teaching, and \fknowers benefit more from seeing examples that target \linearconcept. 
    }
    \label{fig:function_data_adaptation}
\end{figure*}

\begin{figure*}[t!]
    \centering
    \includegraphics[width=0.9\textwidth]{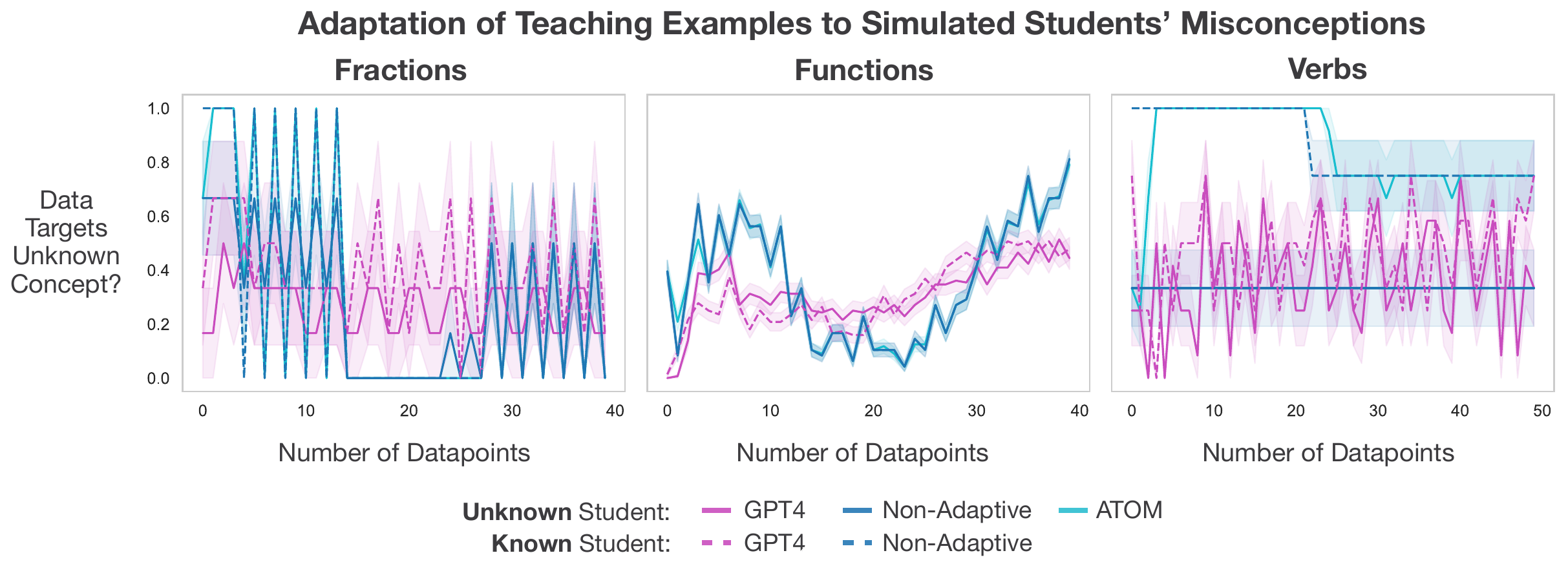}
    \caption{Plot showing whether examples selected by teaching methods target student misconceptions. For fractions, multiplication problems with common denominators target \addgeneralizers' misconceptions, and addition problems with different denominators target \multgeneralizers' misconceptions. For functions, inputs x for which \fx returns True target \gknowers' misconceptions, and inputs for which \fx returns False target \fknowers' misconceptions.  
    For verbs, inputs with class ``unknown'' by the student (\eg a \verbclass{+ied} verb for a \verbclass{+ied-learner}) target the 
    student's misconception.
    }
    \label{fig:synthetic_adaptation_all}
\end{figure*}

For the simulated evaluations, we analyze whether the examples selected by different teaching methods target students' specific misconceptions.

\paragraph{Functions} Figure~\ref{fig:function_data_adaptation} shows whether the selected teaching examples target learning \fnox or \linearconcept in the target \wug concept, split by student type. We expect an adaptive teacher to select examples targeting \fnox for \gknowers at the start of learning, and similarly for examples targeting \linearconcept for \fknowers.

In both plots, we see that the optimal teacher \greedyknownteacher, which assumes access to the ground truth student type, exhibits the expected behavior: It selects more examples targeting \f for the \gknower than for the \fknower at the beginning of the teaching interaction. \adaptiveunknownteacher, despite needing to maintain guesses for the student type, 
shows similar adaptivity to students' priors early on,
selecting more examples targeting \fnox for the \gknower than for the \fknower at the start of teaching.

\gptfourknownteacher also shows this adaptivity when assuming the true student type. However, when it does not have access to the true student type, it does not show this adaptation; the data selection patterns of \gptfourunknownteacher are highly similar for the \fknowers and \gknowers, suggesting that \gptfour struggles with doing implicit adaptation online.

\paragraph{All Tasks} Figure~\ref{fig:synthetic_adaptation_all} shows adaptation of teaching examples across all tasks and student types. We observe similar trends: Across tasks, \adaptiveunknownteacher shows similar levels of adaptivity to \greedyknownteacher, despite not knowing student type, and outperforms both \gptfour methods. We also observe \gptfourknownteacher selecting more examples targeting unknown concepts than \gptfourunknownteacher.

\section{Creating Function Concepts and Student Types}
\label{s:incorrect_concepts}
\gb has possible values $[1, 2, \cdots, 9 ]$ and \ga has possible values $[-5, -4, \cdots, 4, 5]$.
For each concept, to create the \fknower, we uniformly sample the incorrect \gb from the range of possible values of \gb, excluding the target \gb; to create the \gknower, we uniformly sample the incorrect \fnox from a set of spuriously associated functions: These possible values are shown in Table~\ref{tab:function_incorrect_fx}. The full list of concepts and incorrect student beliefs are shown in Table~\ref{tab:function_concepts}.

\subsection{Creating the Verbs Dataset}
\label{s:verbs_appendix}

For the verbs task, we create verb classes with reg exp matching on their past tense forms. For the \gptfour method, we create ground truth outputs by first checking if a lemma exists in the Unimorph dataset; if not, we use a Python verb inflection package \texttt{pyinflect}\footnote{\url{https://github.com/bjascob/pyInflect}} to first get the past tense form of the verb, then categorize it. Table~\ref{tab:verb_classes} shows verb classes and corresponding counts in the resulting dataset.

The model that we fit on the full dataset (to derive the parameters of the target concept) obtains a predictive accuracy of 95.47\%, and the mean probability of the ground truth outputs across the full dataset is 0.945.

\begin{table*}[ht!]
\centering
\small
\begin{tabular}{l@{\hspace{1.25\tabcolsep}}|m{0.25\linewidth}cc}
\toprule
Verb Class & Description & Example & Counts \\ 
\midrule
\verbclass{+ed} & add \emph{`ed'} to the lemma & \emph{clasp} & 6,130 \\
\verbclass{+d} & add \emph{`d'} to the lemma & \emph{smile} & 13,463\\
\verbclass{+ied} & replace last \emph{`y'} with \emph{`ied'} & \emph{cry} & 1,056 \\
\verbclass{+consonant+ed} & double last consonant, add \emph{`ed'} & \emph{stop} & 1,878 \\
\bottomrule
\end{tabular}
\caption{Verb classes and corresponding dataset counts for the verb conjugation task (\sect{s:verb_task}).}
\label{tab:verb_classes}
\end{table*}

\section{Simulated Experiments}
\label{s:appendix_synthetic_learners}

For all tasks, we obtain predictions from simulated students by {sampling} from their predictive distributions.

\paragraph{Program Tasks} For the programmatic tasks (functions, fractions), each simulated student's prior belief in a program \concept is proportional to the number of ``special primitives'' that appear in the program. We derive the prior over programs by multiplying a value $c$ by the number of special primitives that appear in a program to get values for all programs; we then normalize these values to get a distribution over programs.

The simulated students for the function and fraction tasks also maintain a \textbf{noise parameter} that governs how noisy the labels are in the examples they observe; this noise parameter governs their posterior updates. For fractions, this noise parameter is 0.8, and for functions, it is 0.05. We use the same noise values for the teacher's models of the students.

\begin{table*}[t!]
\footnotesize
    \renewcommand{\arraystretch}{0.8}
    \begin{tabular}{p{0.25\textwidth}|p{0.28\textwidth}|p{0.37\textwidth}}
        \toprule
        \multicolumn{3}{l}{Addition Sub-Programs}\\\midrule
        \multicolumn{1}{c}{\textbf{(1)}} & \multicolumn{1}{c}{(2)} & \multicolumn{1}{c}{(3)}\\
        & \codecell{\multirow{4}{*}{add numerators \& denominators}} & \codecell{\ttfamily \footnotesize if denominators are equal:} \\ \codecell{\textbf{make common denominators}} && \quad \codecell{add numerators}\\ \ttfamily \footnotesize \textbf{add numerators} && \ttfamily \footnotesize else:\\ && \ttfamily \footnotesize \quad add numerators \& denominators\\\midrule
        \multicolumn{3}{l}{Multiplication Sub-Programs}\\\midrule
        \multicolumn{1}{c}{(4)} & \multicolumn{1}{c}{\textbf{(5)}} & \multicolumn{1}{c}{(6)}\\
        & \ttfamily \footnotesize \multirow{4}{*}{\textbf{multiply numerators \& denoms}} & \ttfamily \footnotesize if denominators are equal:\\ \ttfamily \footnotesize make common denominators && \quad \codecell{multiply numerators}\\ \ttfamily \footnotesize multiply numerators && \codecell{else:}\\ && \ttfamily \footnotesize \quad multiply numerators \& denominators\\
    \bottomrule
    \end{tabular}
    \caption{
    The addition and multiplication sub-programs in the concept space for the fraction task in \benchmark \sectpar{s:benchmark}. The sub-programs in the target concept (\ie correct sub-programs for adding/multiplying fractions) are \textbf{bolded}.
    }
\label{tab:fraction_concepts}
\end{table*}

\paragraph{Fractions} Table~\ref{tab:fraction_concepts} shows the multiplication and addition sub-programs that are used to create the concept space for simulated students for the fraction task. For the \addgeneralizer who over-generalizes the procedure for addition, the ``special primitives'' are (1), (4), and (6). For the \multgeneralizer who over-generalizes the procedure for multiplication, the ``special primitives'' are (2), (3), and (5). We use a value of $c=1\mathrm{e}{5}$.

\paragraph{Functions}
For \fknowers, the ``special primitives'' are the target \fnox and spurious \gb (and so programs with either of these primitives would have higher prior beliefs; programs having \emph{both} the target \fnox and spurious \gb would have the highest prior belief). Similarly, for \gknowers, the incorrect \fnox and target \gb are the special primitives. We use a value of $c=1\mathrm{e}{4}$.

\begin{figure*}[ht!]
    \centering
    \includegraphics[width=\textwidth]{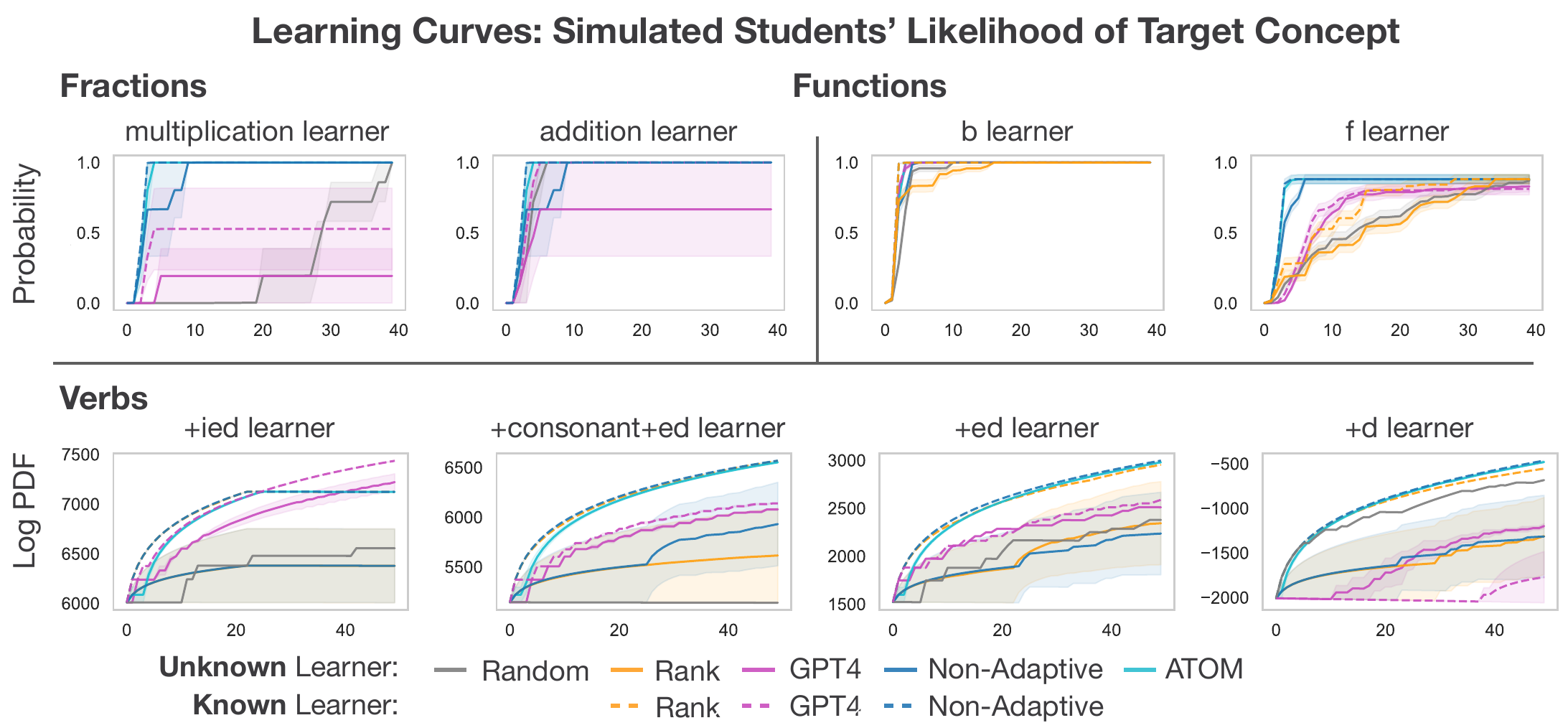}
    \caption{Learning curves for simulated students. Top row: results for function learning and fraction arithmetic, with y-axis showing the probability of the target concept. Bottom row: results for verb conjugation, with y-axis showing the log PDF of the target concept. Each subplot corresponds to a different student type. Color indicates teaching method. Linestyles indicate whether the true student type is assumed (\emph{dashed}=unknown, \emph{solid}=known).}
    \label{fig:synthetic_learning_all_curves}
\end{figure*}

\begin{figure}[b!]
    \centering
    \includegraphics[width=\columnwidth]{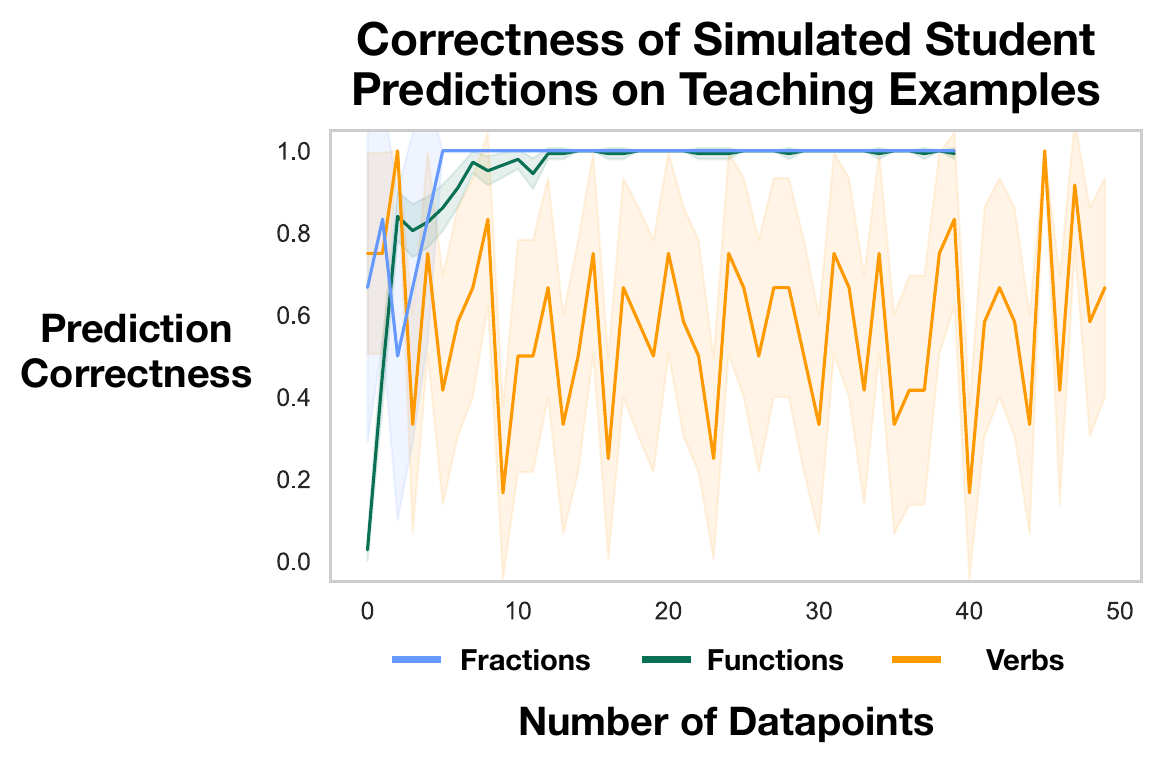}

    \caption{Correctness of simulated student predictions on teaching examples by task. Error bars indicate 95\% confidence intervals.}
    \label{fig:student_acc_teaching_examples}
\end{figure}

\begin{figure*}[h!]
    \centering
    \includegraphics[width=\textwidth]{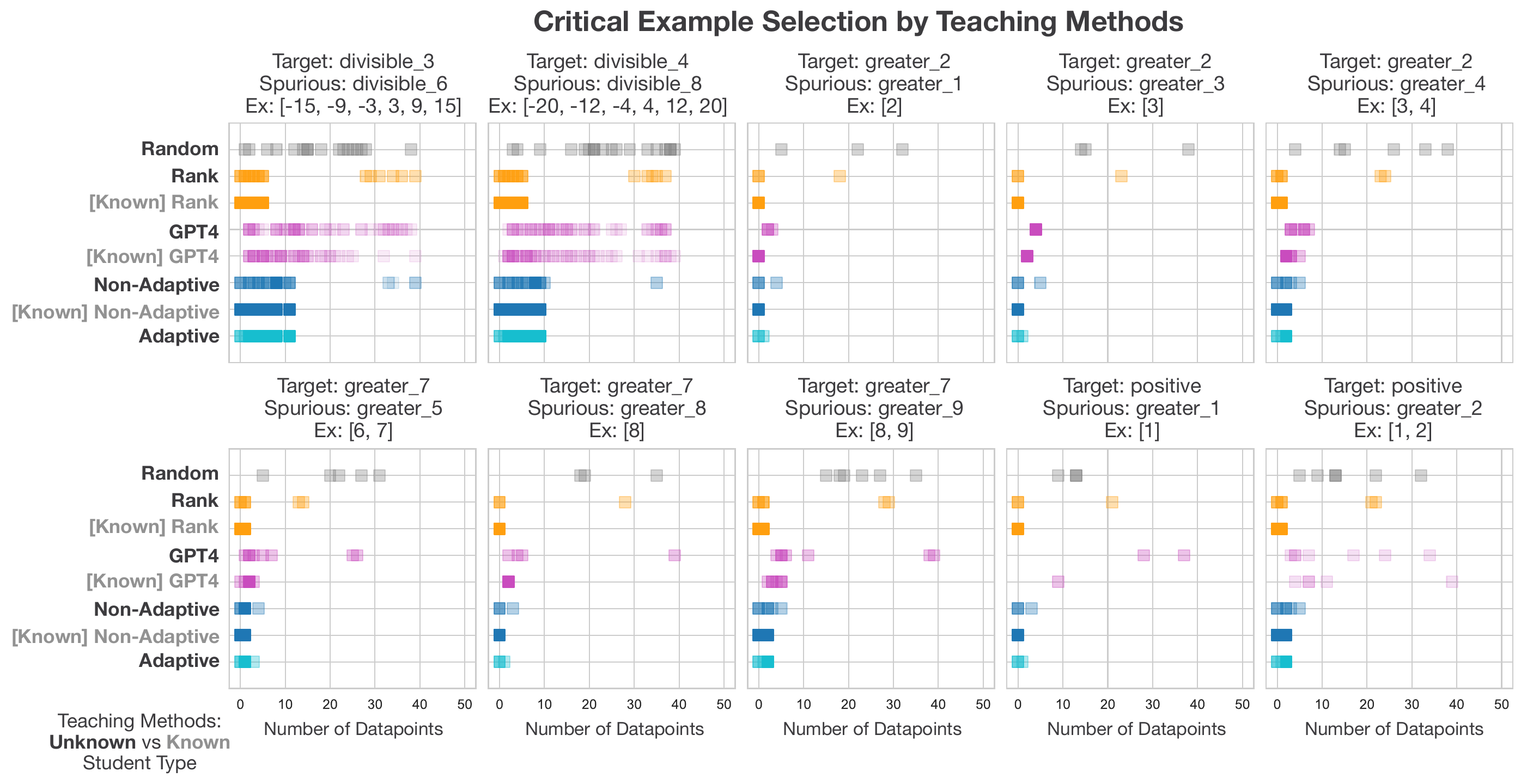}
    \caption{Full results for critical example selection by different teaching methods for the function task. Results are for simulated \gknowers, who have a spurious belief about \fnox that agrees with the target \fnox on all but a few examples, as labeled. The opacity of each square corresponds to the mean value of whether the example chosen by the teaching method at that step in learning is a ``critical example'' (averaged across experimental conditions: seed and concepts). See \sect{s:critical_examples} for details. 
    }
    \label{fig:critical_examples_full}
\end{figure*}

\section{\gptfour: Details}
\label{s:prompts}

Prompts for \gptfour are shown in Tables~\ref{tab:fraction_prompts_known}, \ref{tab:fraction_prompts_unknown}, and \ref{tab:fraction_prompts_student} for fractions, Tables~\ref{tab:function_prompts_known}, \ref{tab:function_prompts_unknown}, and \ref{tab:function_prompts_student} for functions, and Tables~\ref{tab:verb_prompts_known}, \ref{tab:verb_prompts_unknown}, and \ref{tab:verb_prompts_student} for verbs. An example conversation between \gptfourunknownteacher and a simulated student is shown in Table~\ref{tab:gpt_example_conversation}.

\paragraph{Processing/Filtering \gptfour Outputs}

In order to control for the fact that \gptfour may generate incorrect outputs for examples, we use {ground truth} outputs for generated inputs. We parse \gptfour-generated messages to obtain inputs, then compute ground truth labels for those inputs and append them to the message history, starting with ``That's correct/incorrect.'' If a message cannot be parsed, we append a canned response, \ie \emph{``Sorry, I could not learn from that example. I can only learn from examples that are formatted as...''} (if no output can be parsed from the message) or \emph{``I would like to keep learning. Can I have another example?''} (if no input can be parsed from the message); these messages do not count as an ``interaction'' in comparing against other teaching methods. For human experiments, we do not display these canned responses to the students and instead only display messages asking for predictions on examples and providing ground truth answers.

\section{Human Experiments}
\label{s:human_appendix}

\begin{table*}[t!]
\centering
    \begin{tabular}{l}
        \toprule
        \texttt{\thead[l]{Dr. Smith spent a bunch of time studying this machine. She figured out that when wug is defined, it\\ computes a function of the form \linearconcept, where \ga and \gb are constant numbers, so you only need to figure\\ out what \ga and \gb are.\\\\
        She also left a note with some thoughts:\\
        I'm pretty sure, but not totally confident, that:\\
        1) wug is undefined when inputs are \textbf{greater than 2}\\
        2) When wug is defined, \textbf{b = 3}\\
        --Dr. Smith\\\\
        Dr. Smith is quite familiar with wug, so her note should give you a good place to start! But keep in\\ mind that it is possible that she is wrong.
        }}\\
    \bottomrule
    \end{tabular}
    \caption{An example of a hint given to a human learner. (1) is correct, while (2) is not, thus creating an \fknower.}
    \label{tab:human_hints}
    \end{table*}

\subsection{Set-Up}

\paragraph{Post-Processing} We filter and rerun any experiments where the chat messages were logged out of order or sent twice.

\paragraph{Hyperparameters} The \adaptiveunknownteacher uses a noise parameter value of 0.02 for modeling simulated students.

\begin{figure*}
\centering

\includegraphics[page=1,width=0.8\linewidth]{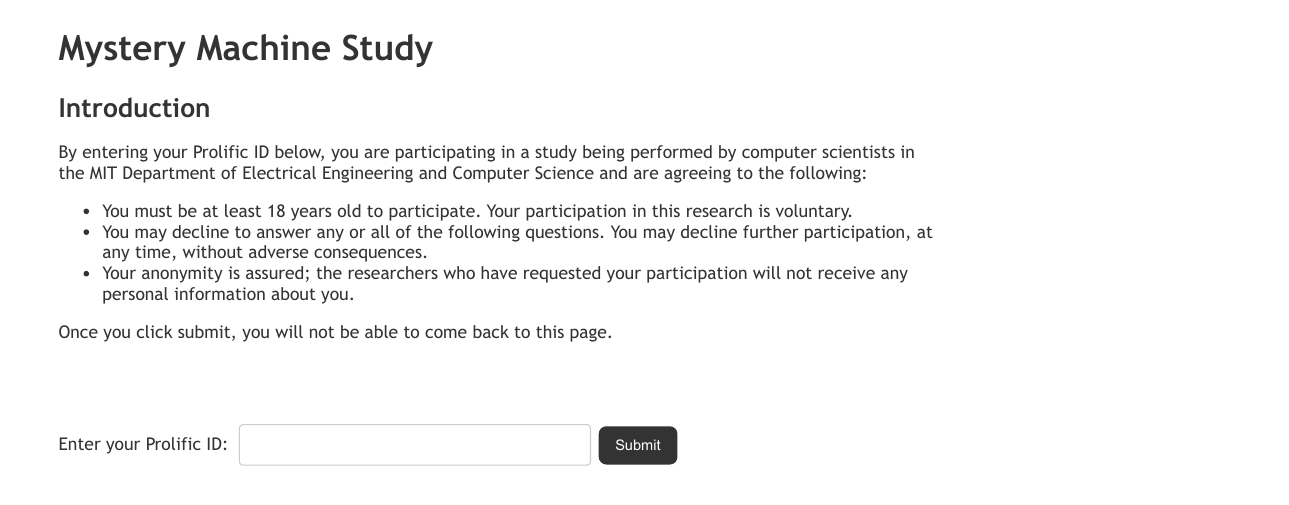}
\includegraphics[page=2,width=0.8\linewidth]{appendix/figures/instructions.pdf}
\includegraphics[page=3,width=0.8\linewidth]{appendix/figures/instructions.pdf}

\caption{Screens 1-3 (instructions) for the study with human participants.}
\label{fig:human_1}
\end{figure*}

\begin{figure*}
\centering

\includegraphics[page=4,width=0.8\linewidth]{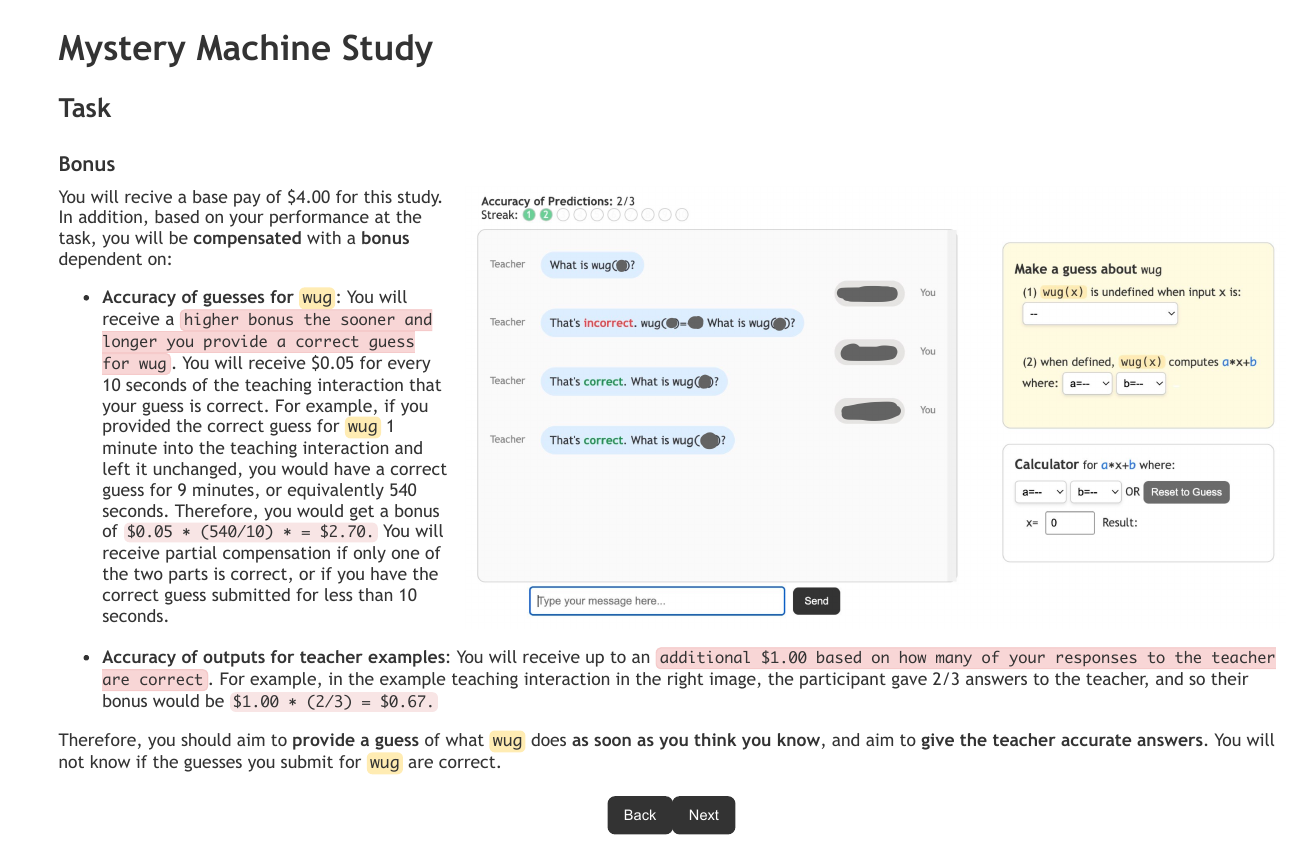}
\includegraphics[page=5,width=0.8\linewidth]{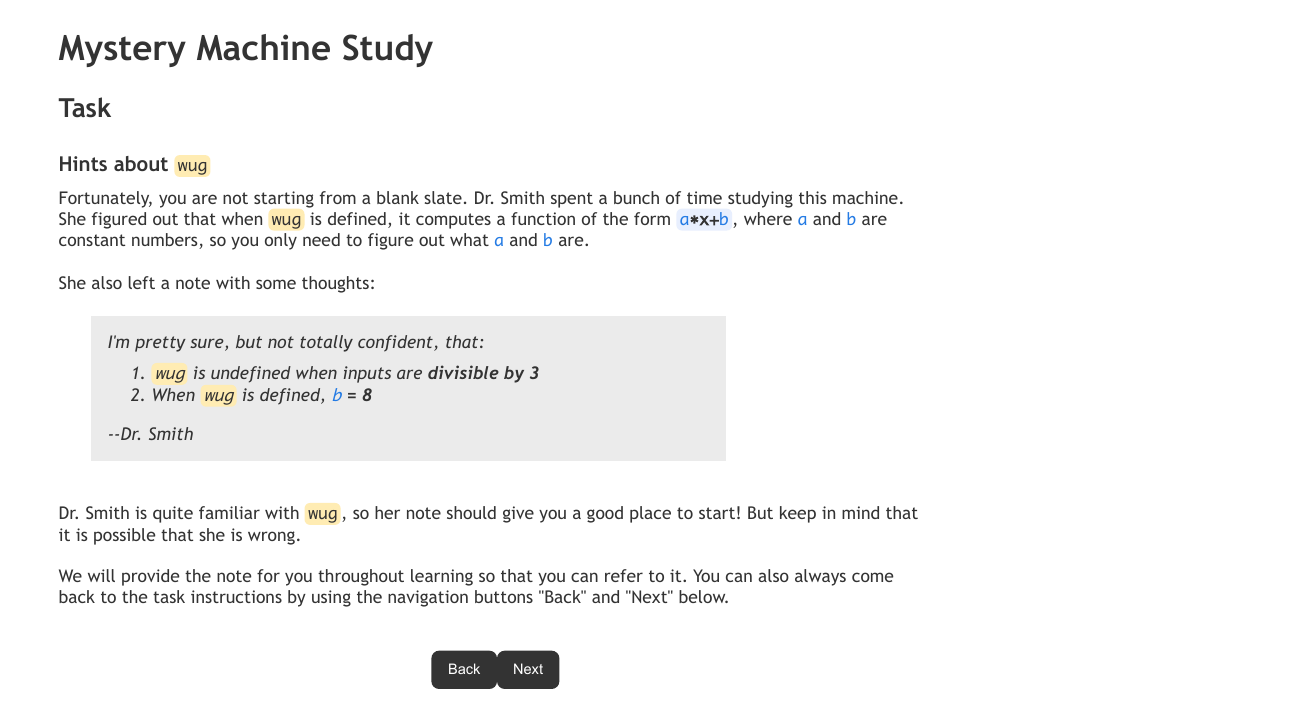}
\includegraphics[page=6,width=0.8\linewidth]{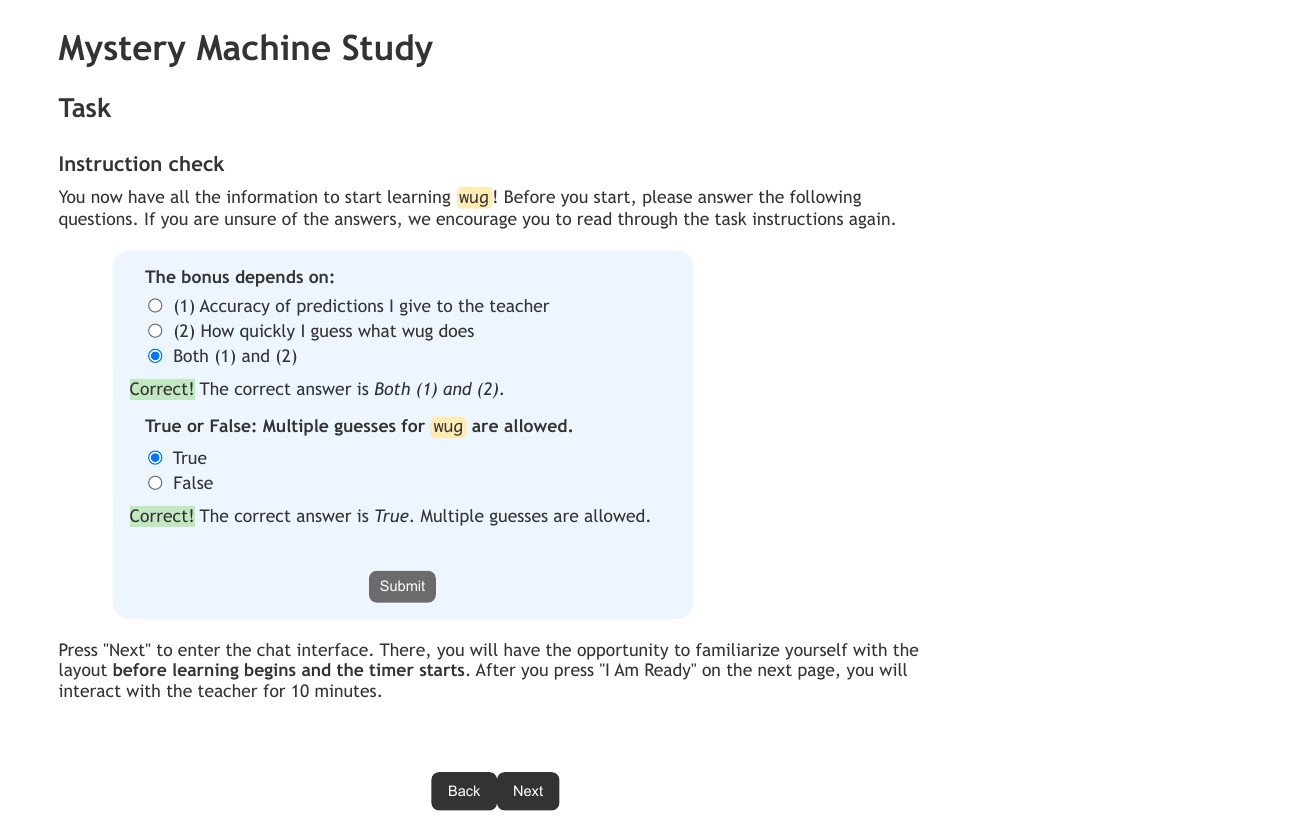}

\caption{Screens 4-6 (instructions) for the study with human participants.}
\label{fig:human_2}
\end{figure*}

\begin{figure*}
\centering

\includegraphics[page=1,width=0.8\linewidth]{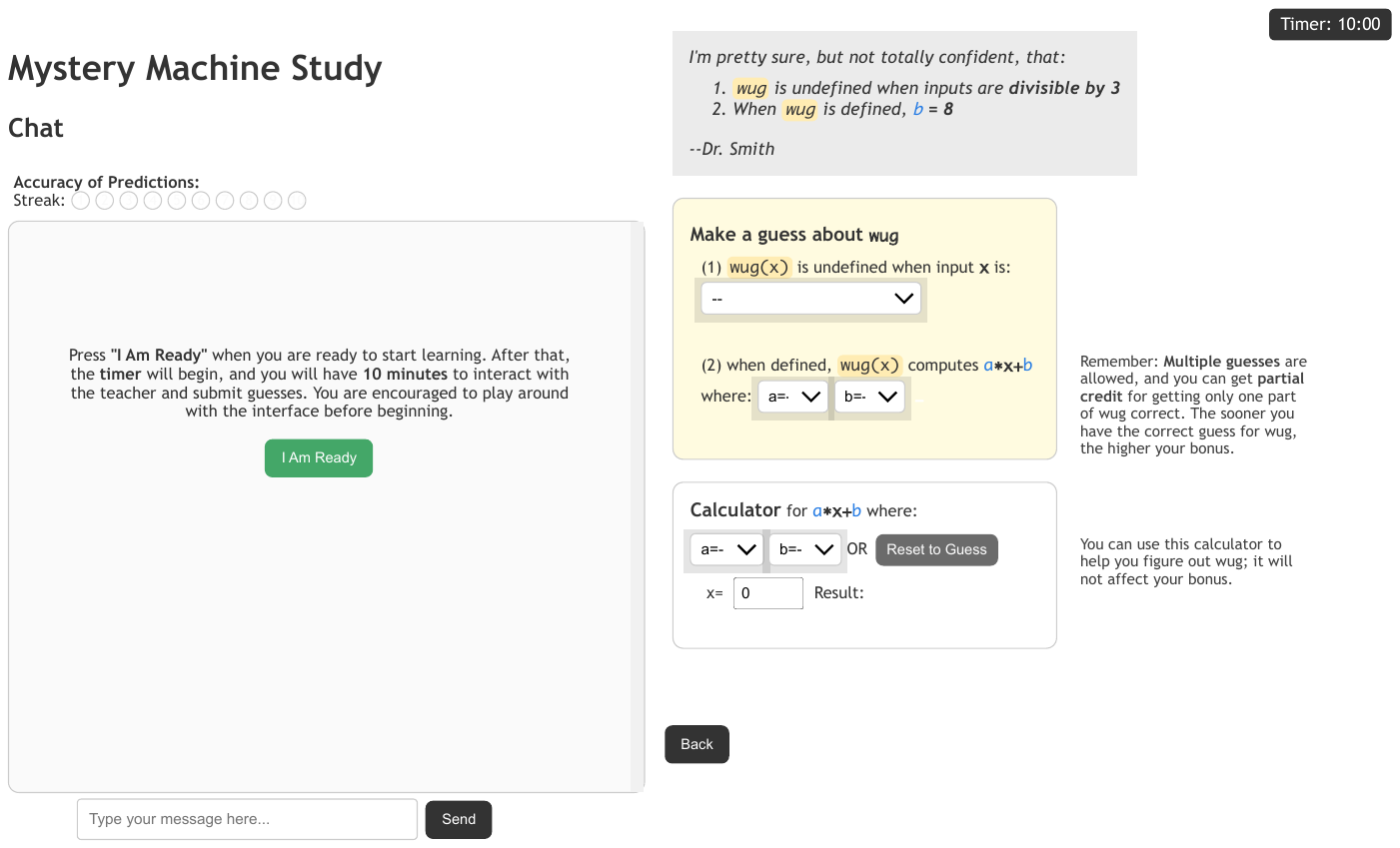}
\includegraphics[page=2,width=0.8\linewidth]{appendix/figures/secondhalf.pdf}

\caption{The chat interface for the study with human participants.}
\label{fig:human_3}
\end{figure*}

\begin{figure*}
\centering

\includegraphics[page=3,width=0.8\linewidth]{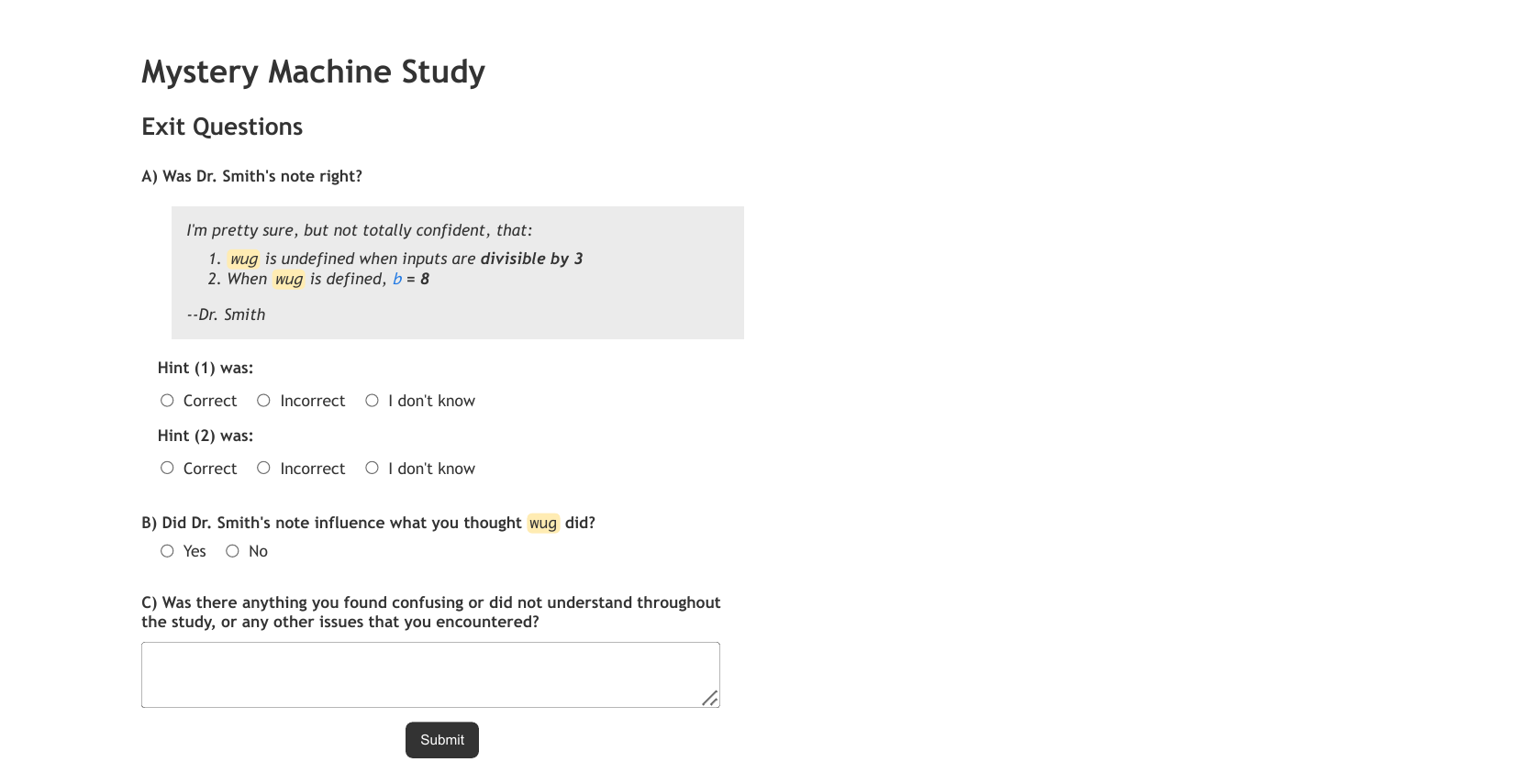}
\includegraphics[page=4,width=0.8\linewidth]{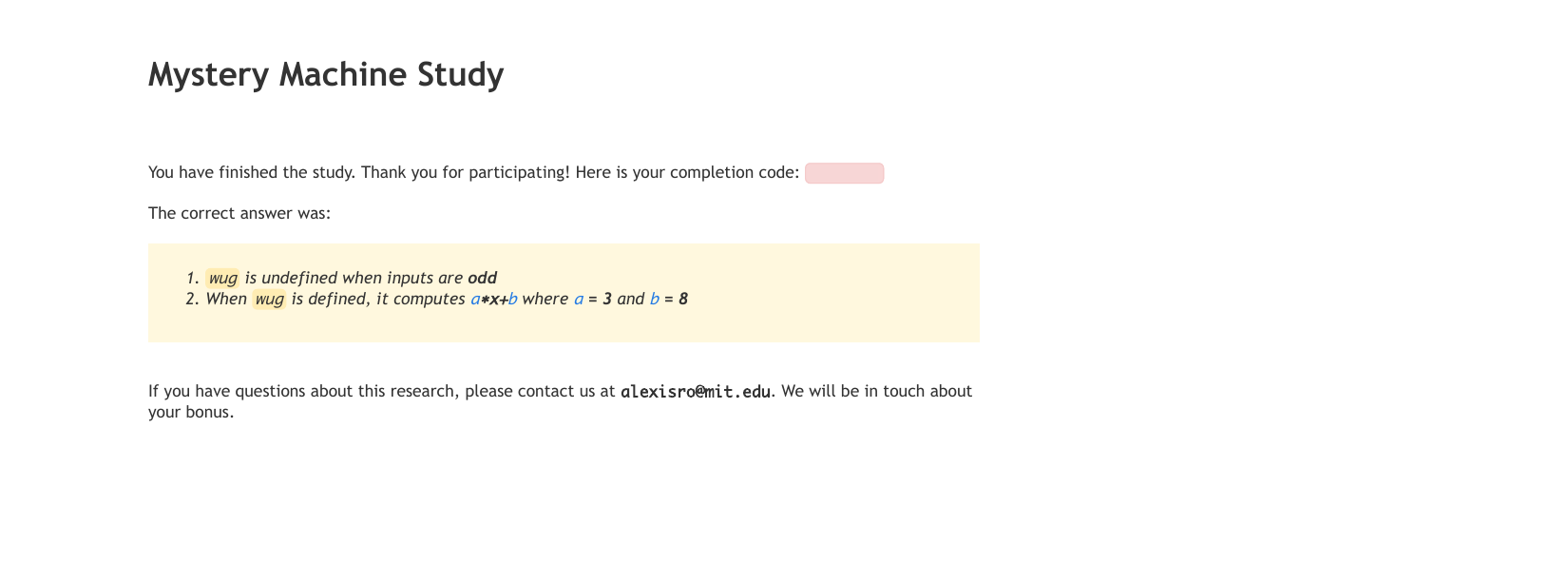}

\caption{Post-chat end screens for the study with human participants.}
\label{fig:human_4}
\end{figure*}

\paragraph{Instructions} An example of hints given to the human participants is shown in Table~\ref{tab:human_hints}. The full set of instructions shown to human participants, along with the interface, are shown in Figures~\ref{fig:human_1}/\ref{fig:human_2} (instructions), \ref{fig:human_3} (chat), and \ref{fig:human_4} (end).

\paragraph{Bonus Compensation} Participants are told that their bonus depends on two things:

\begin{enumerate}
    \item Accuracy of \wug guesses: Participants are told they will receive $0.05$ for every 10 seconds of the teaching interaction that their guess is correct, with partial compensation if only \fnox or only \ga/\gb is correct.
    \item Accuracy of predictions on teaching examples: Participants are told they will receive up to an additional $1.00$ based on the accuracy of their predictions.
\end{enumerate}

As shown in Figure~\ref{fig:human_2}, participants are prompted to indicate their understanding of what their bonus depends on.

\begin{figure*}[b!]
    \centering
    \includegraphics[width=\columnwidth]{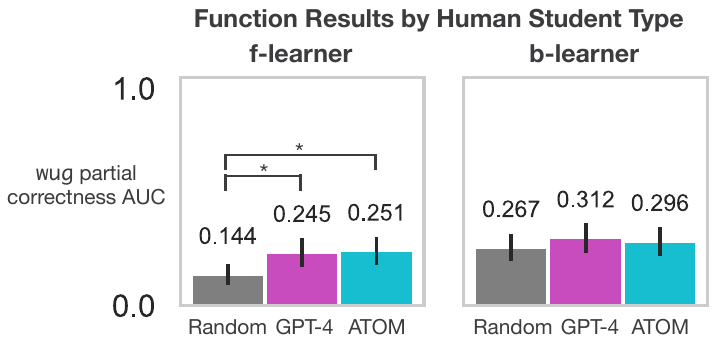}
    \caption{Results showing how efficiently human students guessed the correct \wug concept \sectpar{s:human_learning_efficiency}, by student type. Stars indicate statistically significant results under a paired t-test: For \gknowers, $p < 0.05$ for \randomteacher v.s. \adaptiveunknownteacher and for \randomteacher v.s. \gptfourunknownteacher.
    Error bars show 95\% confidence intervals.}
    \label{fig:human_bar_by_learner}
\end{figure*}

\begin{figure*}[ht!]
    \centering
    \includegraphics[width=\textwidth]{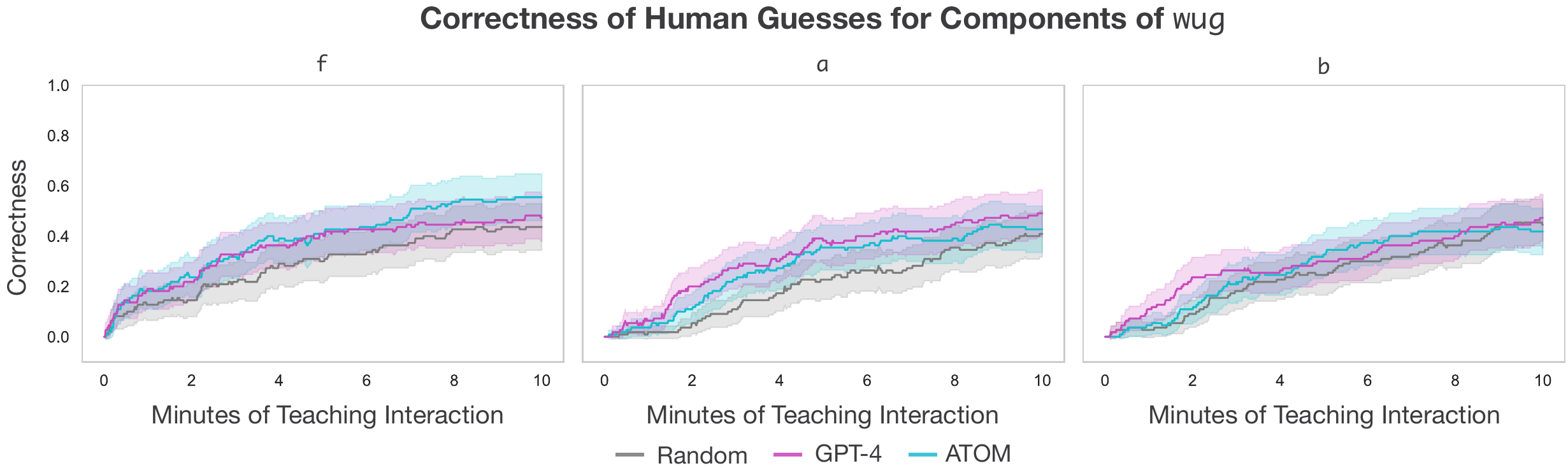}
    \caption{Correctness of humans' guesses of \wug, split by components and as a function of time spent interacting with the teacher. Error bars show 95\% confidence intervals.
    }
    \label{fig:human_learning}
\end{figure*}

\begin{figure*}[ht!]
    \centering
    \includegraphics[width=\textwidth]{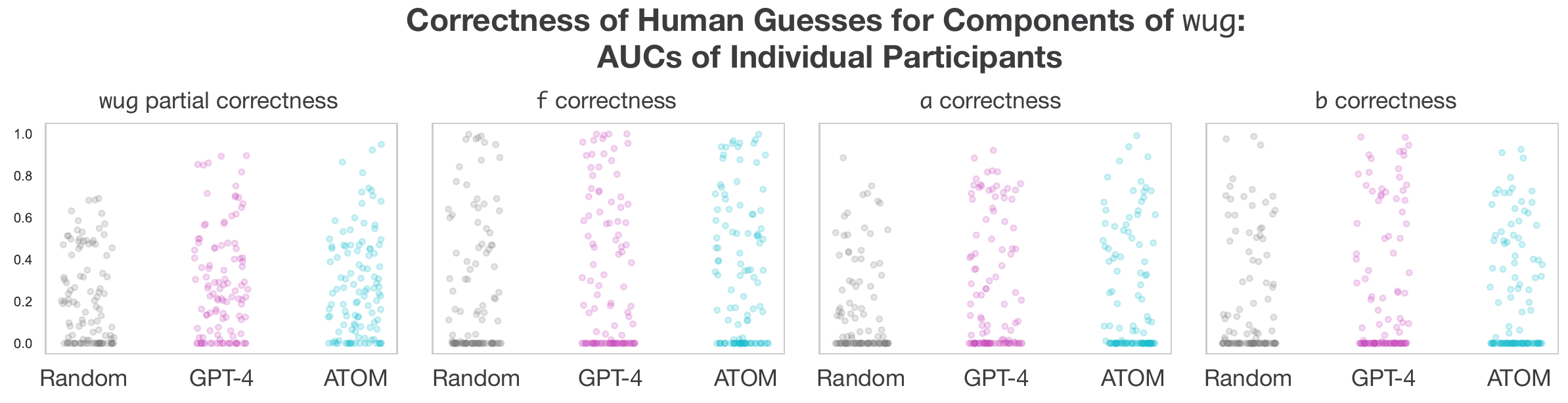}
    \caption{Correctness of guesses of \wug for individual human participants. Each dot shows the AUC of the curve of the metric's correctness over time for an individual participant.
    }
    \label{fig:human_scatter}
\end{figure*}

\begin{figure*}[b!]
    \centering
    \includegraphics[width=\columnwidth]{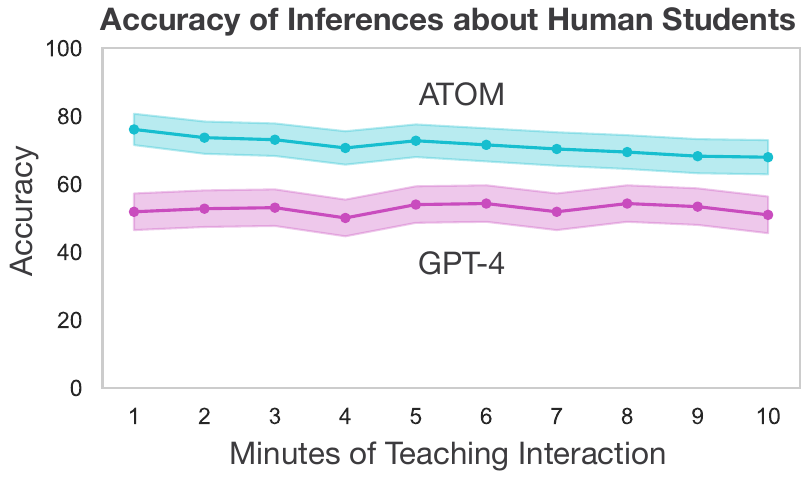}
    \caption{Accuracy of teaching methods' predictions of student type with human learners (\sect{ssec:human_student_accuracy}). Error bars show 95\% confidence intervals. 
    }
    \label{fig:human_student_accuracy}
\end{figure*}

\subsection{Additional Results}
\label{s:human_appendix_results}
Figure~\ref{fig:human_bar_by_learner} shows the learning efficiency of human students by student type. Figure~\ref{fig:human_learning} shows the correctness of \wug guesses by component as a function of time. Figure~\ref{fig:human_scatter} shows the AUCs of \wug correctness by component for individual participants. Figure ~\ref{fig:human_student_accuracy} shows the accuracy of student type predictions over time.

\subsection{Combining \gptfour and \adaptiveunknownteacher} 
\label{s:gpt_atom_combination}
We run an experiment combining \gptfour and \adaptiveunknownteacher in the following way: We use \adaptiveunknownteacher to make inferences about student type, then prompt \gptfour with \adaptiveunknownteacher's inference by updating the system prompt to \gptfour after each prediction made by the student. Before any predictions are given, \gptfour is prompted with both student types (\ie with the prompt given to the teacher that does not know student type). We call this method \combinedteacher.

\begin{figure*}[b!]
    \centering
    \includegraphics[width=0.75\columnwidth]{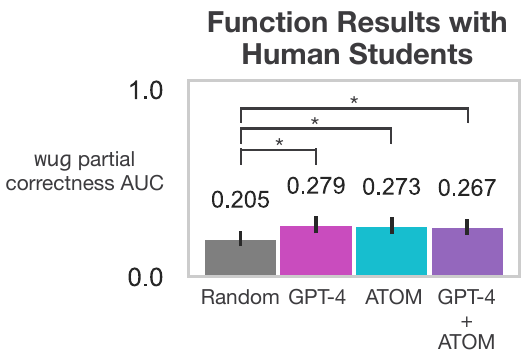}
    \caption{Results with human students showing how efficiently students guessed the correct \wug concept \sectpar{s:human_learning_efficiency}. Stars indicate statistically significant results under a paired t-test. Error bars show 95\% confidence intervals.}
    \label{fig:human_bar_combo}
\end{figure*}

Results are shown in Figure~\ref{fig:human_bar_combo}. We find that \combinedteacher outperforms \randomteacher ($p < 0.05$ using a paired t-test) but does not outperform \gptfourunknownteacher or \adaptiveunknownteacher.

\begin{table*}[t]
\centering
\small
\begin{tabular}{lll|ll|c}
\toprule
\multicolumn{3}{c}{Target} & \multicolumn{2}{c}{Incorrect} & Used in \\
\fnox & \ga & \gb & \fnox & \gb & Human Experiments? \\\midrule\midrule
\f{even} & 1 & 7 & \f{divisible\_6} & 5 \\
\f{even} & -5 & 5 & \f{divisible\_6} & 7 & Y\\
\f{even} & 3 & 8 & \f{divisible\_4} & 3 & Y\\
\f{greater\_2} & 1 & 7 & \f{greater\_4} & 3 \\
\f{greater\_2} & -5 & 5 & \f{greater\_1} & 6 \\
\f{greater\_2} & 3 & 8 & \f{greater\_3} & 5 & Y\\
\f{prime} & 1 & 7 & \f{odd} & 2 & Y\\
\f{prime} & -5 & 5 & \f{odd} & 9 & Y\\
\f{prime} & 3 & 8 & \f{odd} & 6 \\
\f{divisible\_3} & 1 & 7 & \f{divisible\_6} & 6 \\
\f{divisible\_3} & -5 & 5 & \f{divisible\_6} & 9 \\
\f{divisible\_3} & 3 & 8 & \f{divisible\_6} & 5 \\
\f{divisible\_4} & 1 & 7 & \f{divisible\_8} & 1 \\
\f{divisible\_4} & -5 & 5 & \f{divisible\_8} & 8 & Y\\
\f{divisible\_4} & 3 & 8 & \f{divisible\_8} & 9 & Y \\
\f{positive} & 1 & 7 & \f{greater\_2} & 4 \\
\f{positive} & -5 & 5 & \f{greater\_2} & 2 \\
\f{positive} & 3 & 8 & \f{greater\_1} & 4 \\
\f{odd} & 1 & 7 & \f{divisible\_5} & 3 & Y\\
\f{odd} & -5 & 5 & \f{prime} & 2 & Y\\
\f{odd} & 3 & 8 & \f{divisible\_3} & 6 & Y\\
\f{greater\_7} & 1 & 7 & \f{greater\_9} & 2 \\
\f{greater\_7} & -5 & 5 & \f{greater\_8} & 6 & Y\\
\f{greater\_7} & 3 & 8 & \f{greater\_5} & 6 \\
\bottomrule
\end{tabular}
\caption{The target concepts and incorrect beliefs used in the function learning experiments. The \textbf{target} \fnox, \ga, \gb define the concepts being taught. The \textbf{incorrect} \fnox is the belief that the \gknowers have about \fnox at the start of learning, and the incorrect \gb is the value that the \fknowers believes \gb to be.}
\label{tab:function_concepts}
\end{table*}

\begin{table*}[]
\centering
\small
\begin{tabular}{l@{\hspace{1.25\tabcolsep}}|p{0.75\textwidth}}
\toprule
\fnox & Incorrect \fnox Options \\ 
\midrule
\f{prime} & \f{odd} \\
\f{positive} & \f{greater\_n} for $n \in [-2, -1, 1, 2]$\\
\f{even} & \f{divis\_4}, \f{divis\_6} \\
\f{odd} & \f{prime}, \f{divis\_3}, \f{divis\_5}, \f{divis\_7}\\
\f{divis\_{n}} & \f{divis\_m} where $m$ is the smallest multiple of $n$ or the largest factor of $n$ (if $m=2$, this is \f{even}) \\
\f{greater\_{n}} & \f{greater\_{m}} where $\abs{m-n}<=2$ (if $m=0$, this is \f{positive}) \\
\bottomrule
\end{tabular}
\caption{Descriptions of how the options for incorrect \fnox beliefs of \gknowers are generated for each target \fnox in the function task. We uniformly sample the incorrect \fnox from the set of options to determine the actual incorrect \fnox belief for the \gknowers.}
\label{tab:function_incorrect_fx}
\end{table*}

\begin{table*}[t!]
    \renewcommand{\arraystretch}{0.8}
    \begin{tabular}{p{\textwidth}}
        \toprule
        \ttfamily \footnotesize You are GPT-teacher, an expert teacher. Your goal is to teach a student how to multiply and add fractions as efficiently as possible with helpful examples.\\ \ttfamily \footnotesize \\ \ttfamily \footnotesize You will be interacting with a student who has spent some time with fraction arithmetic but still has some misconceptions about how it works. The student you will be interacting with is a student who \textbf{performs multiplication correctly, but tends to incorrectly add both numerators and denominators when adding fractions, especially when denominators are different}. \\ \ttfamily \footnotesize \\ \ttfamily \footnotesize Please make sure to follow these instructions:\\ \ttfamily \footnotesize - You are only allowed to give students example fraction problems, and ask them to guess the outputs. You may not explain any concepts to them directly, or ask any other questions. Anything other than example fraction problems and answers will be ignored by the student.\\ \ttfamily \footnotesize - The student has not learned how to simplify fractions yet, so please do not simplify the fractions in your examples. Leave the answers in their unsimplified form. The student will also not simplify their answer.\\ \ttfamily \footnotesize - Please only use fractions with positive numerators and denominators.\\ \ttfamily \footnotesize - Do not teach arithmetic with mixed numbers or whole numbers.\\ \ttfamily \footnotesize - Only teach fraction addition and multiplication. Please format input/output examples as: a/b+c/d=e/f for addition or a/b*c/d=e/f for multiplication.\\ \ttfamily \footnotesize - Keep teaching with fraction problems and outputs until the student says they would like to stop, even if you think you have covered the full input range.\\ \ttfamily \footnotesize \\ \ttfamily \footnotesize For example, your interactions will look like the following, where capital words indicate placeholders for actual verb lemmas and categories:\\ \ttfamily \footnotesize \\ \ttfamily \footnotesize Your interactions will look like the following (where letters are placeholders for actual numbers): \\ \ttfamily \footnotesize System: What is a/b+c/d?\\ \ttfamily \footnotesize User: a/b+c/d=e/f \\ \ttfamily \footnotesize System: That's [correct/incorrect]. a/b+c/d=x/y. What is g/h+i/j? \\ \ttfamily \footnotesize \\ \ttfamily \footnotesize Please start by asking the student for their guess on a fraction example.\\
    \bottomrule
    \end{tabular}
    \caption{
        System prompt to \gptfourknownteacher for the fraction task (known student type). \textbf{Bolded} words indicate variables that change between student types.
    }
    \label{tab:fraction_prompts_known}
\end{table*}

\begin{table*}[t!]
    \renewcommand{\arraystretch}{0.8}
    \begin{tabular}{p{\textwidth}}
        \toprule
        \ttfamily \footnotesize You are GPT-teacher, an expert teacher. Your goal is to teach a student how to multiply and add fractions as efficiently as possible with helpful examples.\\ \ttfamily \footnotesize \\ \ttfamily \footnotesize You will be interacting with a student who has spent some time with fraction arithmetic but still has some misconceptions about how it works. There are 2 kinds of students: \\ \ttfamily \footnotesize 1) Students who perform addition correctly, but tend to incorrectly multiply only numerators when multiplying fractions, especially when the denominators are equal; if the denominators are not equal, the student sometimes makes common denominators and then multiplies the numerators\\ \ttfamily \footnotesize 2) Students who perform multiplication correctly, but tend to incorrectly add both numerators and denominators when adding fractions, especially when denominators are different\\ \ttfamily \footnotesize You should try to figure out which kind of student you are interacting with and then teach them accordingly.\\ \ttfamily \footnotesize \\ \ttfamily \footnotesize Please make sure to follow these instructions:\\ \ttfamily \footnotesize - You are only allowed to give students example fraction problems, and ask them to guess the outputs. You may not explain any concepts to them directly, or ask any other questions. Anything other than example fraction problems and answers will be ignored by the student.\\ \ttfamily \footnotesize - The student has not learned how to simplify fractions yet, so please do not simplify the fractions in your examples. Leave the answers in their unsimplified form. The student will also not simplify their answer.\\ \ttfamily \footnotesize - Please only use fractions with positive numerators and denominators.\\ \ttfamily \footnotesize - Do not teach arithmetic with mixed numbers or whole numbers.\\ \ttfamily \footnotesize - Only teach fraction addition and multiplication. Please format input/output examples as: a/b+c/d=e/f for addition or a/b*c/d=e/f for multiplication.\\ \ttfamily \footnotesize - Keep teaching with fraction problems and outputs until the student says they would like to stop, even if you think you have covered the full input range.\\ \ttfamily \footnotesize \\ \ttfamily \footnotesize For example, your interactions will look like the following, where capital words indicate placeholders for actual verb lemmas and categories:\\ \ttfamily \footnotesize \\ \ttfamily \footnotesize Your interactions will look like the following (where letters are placeholders for actual numbers): \\ \ttfamily \footnotesize System: What is a/b+c/d?\\ \ttfamily \footnotesize User: a/b+c/d=e/f \\ \ttfamily \footnotesize System: That's [correct/incorrect]. a/b+c/d=x/y. What is g/h+i/j? \\ \ttfamily \footnotesize \\ \ttfamily \footnotesize Please start by asking the student for their guess on a fraction example.\\
    \bottomrule
    \end{tabular}
    \caption{
        System prompt to \gptfourunknownteacher for the fraction task (unknown student type).
    }
    \label{tab:fraction_prompts_unknown}
\end{table*}

\begin{table*}[t!]
    \renewcommand{\arraystretch}{0.8}
    \begin{tabular}{p{\textwidth}}
        \toprule
        \ttfamily \footnotesize Based on this interaction, which kind of student do you think I was at the start of this teaching session:\\ \ttfamily \footnotesize 1) Students who perform addition correctly, but tend to incorrectly multiply only numerators when multiplying fractions, especially when the denominators are equal; if the denominators are not equal, the student sometimes makes common denominators and then multiplies the numerators\\ \ttfamily \footnotesize 2) Students who perform multiplication correctly, but tend to incorrectly add both numerators and denominators when adding fractions, especially when denominators are different\\ \ttfamily \footnotesize \\ \ttfamily \footnotesize Please select (1) or (2).\\
    \bottomrule
    \end{tabular}
    \caption{
        System prompt to \gptfourunknownteacher for inferring student type for the fraction task.
    }
    \label{tab:fraction_prompts_student}
\end{table*}

\begin{table*}[t!]
    \renewcommand{\arraystretch}{0.8}
    \begin{tabular}{p{\textwidth}}
        \toprule
        \ttfamily \footnotesize You are GPT-teacher, an expert teacher. Your goal is to teach a student what a mystery machine called wug does. This machine takes in numbers and outputs numbers. However, it only works for some numbers and is undefined for others. Your goal is to teach the student on what inputs wug is undefined, and when it is defined, what it does. You should do so as efficiently as possible with helpful input/output examples, such as edge cases. \\ \ttfamily \footnotesize \\ \ttfamily \footnotesize The wug machine works as follows: wug(x) is undefined when x is \textbf{greater than 2}. When defined, wug(x) computes \textbf{x+7}.\\ \ttfamily \footnotesize \\ \ttfamily \footnotesize You're going to be interacting with a student who is learning how wug works. The student knows that wug is sometimes undefined. The student also knows that when wug is defined, it computes something of the form a*x+b. In the real wug machine, \textbf{a=1} and \textbf{b=7}. However, the student does not know this. The student only knows that a is a constant number between -5 and 5 (inclusive) and that b is a constant number between 1 and 9 (inclusive).\\ \ttfamily \footnotesize \\ \ttfamily \footnotesize The student knows that wug is undefined when the input is one of the following:\\ \ttfamily \footnotesize - prime\\ \ttfamily \footnotesize - positive\\ \ttfamily \footnotesize - even\\ \ttfamily \footnotesize - odd\\ \ttfamily \footnotesize - divisible by n for n between 3 and 20 (inclusive)\\ \ttfamily \footnotesize - greater than n for n between 1 and 20 (inclusive)\\ \ttfamily \footnotesize \\ \ttfamily \footnotesize Students have varying previous exposure to wug, and so they understand different parts of how wug works. The student you will be interacting with is a student who correctly thinks that \textbf{b=7} but incorrectly thinks that wug is undefined when inputs are \textbf{greater than 4}. \\ \ttfamily \footnotesize \\ \ttfamily \footnotesize Please make sure to follow these instructions:\\ \ttfamily \footnotesize - You are only allowed to give students example inputs, and ask them to guess outputs. You may not explain aspects of the concept to them directly, or ask any other questions. Anything other than inputs and outputs will be ignored by the student.\\ \ttfamily \footnotesize - Please format input/output examples as: wug(INPUT)=ANSWER\\ \ttfamily \footnotesize - wug only works for numbers between -20 to 20 (inclusive), so restrict the inputs you choose to that range. Any inputs outside of that range will be ignored by the student.\\ \ttfamily \footnotesize - Keep teaching with inputs and outputs until the student says they would like to stop, even if you think you have covered the full input range.\\ \ttfamily \footnotesize \\ \ttfamily \footnotesize For example, your interactions will look like the following, where capital words indicate placeholders for actual numbers:\\ \ttfamily \footnotesize \\ \ttfamily \footnotesize Your interactions will look like the following: \\ \ttfamily \footnotesize System: What is wug(INPUT)?\\ \ttfamily \footnotesize User: wug(INPUT)=GUESS\\ \ttfamily \footnotesize System: That's [correct/incorrect]. wug(INPUT)=ANSWER. What is wug(NEW INPUT)?\\ \ttfamily \footnotesize \\ \ttfamily \footnotesize Please start by asking the student for their guess on an input.\\
    \bottomrule
    \end{tabular}
    \caption{
        System prompt to \gptfourknownteacher for the function task (known student type). \textbf{Bolded} words indicate variables that change between student types and target concepts.
    }
    \label{tab:function_prompts_known}
\end{table*}

\begin{table*}[t!]
    \renewcommand{\arraystretch}{0.8}
    \begin{tabular}{p{\textwidth}}
        \toprule
        \ttfamily \footnotesize You are GPT-teacher, an expert teacher. Your goal is to teach a student what a mystery machine called wug does. This machine takes in numbers and outputs numbers. However, it only works for some numbers and is undefined for others. Your goal is to teach the student on what inputs wug is undefined, and when it is defined, what it does. You should do so as efficiently as possible with helpful input/output examples, such as edge cases. \\ \ttfamily \footnotesize \\ \ttfamily \footnotesize The wug machine works as follows: wug(x) is undefined when x is \textbf{greater than 2}. When defined, wug(x) computes \textbf{x+7}.\\ \ttfamily \footnotesize \\ \ttfamily \footnotesize You're going to be interacting with a student who is learning how wug works. The student knows that wug is sometimes undefined. The student also knows that when wug is defined, it computes something of the form a*x+b. In the real wug machine, a=1 and b=7. However, the student does not know this. The student only knows that a is a constant number between -5 and 5 (inclusive) and that b is a constant number between 1 and 9 (inclusive).\\ \ttfamily \footnotesize \\ \ttfamily \footnotesize The student knows that wug is undefined when the input is one of the following:\\ \ttfamily \footnotesize - prime\\ \ttfamily \footnotesize - positive\\ \ttfamily \footnotesize - even\\ \ttfamily \footnotesize - odd\\ \ttfamily \footnotesize - divisible by n for n between 3 and 20 (inclusive)\\ \ttfamily \footnotesize - greater than n for n between 1 and 20 (inclusive)\\ \ttfamily \footnotesize \\ \ttfamily \footnotesize Students have varying previous exposure to wug, and so they understand different parts of how wug works. There are two kinds of students:\\ \ttfamily \footnotesize 1) Students who correctly think that \textbf{b=7} but incorrectly think wug is undefined when inputs are \textbf{greater than 4}\\ \ttfamily \footnotesize 2) Students who correctly think that wug is undefined when inputs are \textbf{greater than 2} but incorrectly think that \textbf{b=3}\\ \ttfamily \footnotesize \\ \ttfamily \footnotesize Please make sure to follow these instructions:\\ \ttfamily \footnotesize - You are only allowed to give students example inputs, and ask them to guess outputs. You may not explain aspects of the concept to them directly, or ask any other questions. Anything other than inputs and outputs will be ignored by the student.\\ \ttfamily \footnotesize - Please format input/output examples as: wug(INPUT)=ANSWER\\ \ttfamily \footnotesize - wug is only defined for numbers between -20 to 20 (inclusive), so restrict the inputs you choose to that range.\\ \ttfamily \footnotesize - Keep teaching with inputs and outputs until the student says they would like to stop, even if you think you have covered the full input range.\\ \ttfamily \footnotesize \\ \ttfamily \footnotesize For example, your interactions will look like the following, where capital words indicate placeholders for actual numbers:\\ \ttfamily \footnotesize \\ \ttfamily \footnotesize Your interactions will look like the following: \\ \ttfamily \footnotesize System: What is wug(INPUT)?\\ \ttfamily \footnotesize User: wug(INPUT)=GUESS\\ \ttfamily \footnotesize System: That's [correct/incorrect]. wug(INPUT)=ANSWER. What is wug(NEW INPUT)?\\ \ttfamily \footnotesize \\ \ttfamily \footnotesize Please start by asking the student for their guess on an input.\\
    \bottomrule
    \end{tabular}
    \caption{
        System prompt to \gptfourunknownteacher for the function task (unknown student type). \textbf{Bolded} words indicate variables that change between student types and target concepts.
    }
    \label{tab:function_prompts_unknown}
\end{table*}

\begin{table*}[t!]
    \renewcommand{\arraystretch}{0.8}
    \begin{tabular}{p{\textwidth}}
        \toprule
        \ttfamily \footnotesize Based on this interaction, which kind of student do you think I was at the start of this teaching session:\\ \ttfamily \footnotesize 1) Students who correctly think that \textbf{b=7} but incorrectly think wug is undefined when inputs are \textbf{greater than 4}\\ \ttfamily \footnotesize 2) Students who correctly think that wug is undefined when inputs are \textbf{greater than 2} but incorrectly think that \textbf{b=3}\\ \ttfamily \footnotesize \\ \ttfamily \footnotesize Please select (1) or (2).\\
    \bottomrule
    \end{tabular}
    \caption{
        System prompt to \gptfourunknownteacher for inferring student type for the function task. \textbf{Bolded} words indicate variables that change between student types and target concepts.
    }
    \label{tab:function_prompts_student}
\end{table*}

\begin{table*}[t!]
    \renewcommand{\arraystretch}{0.8}
    \begin{tabular}{p{\textwidth}}
        \toprule
        \ttfamily \footnotesize You are GPT-teacher, an expert teacher. Your goal is to teach a student how to conjugate English past tense verbs as efficiently as possible with helpful examples.\\ \ttfamily \footnotesize \\ \ttfamily \footnotesize Specifically, your goal is to teach students about four categories of past tense verbs:\\ \ttfamily \footnotesize - `+ed': add `ed' to the verb lemma\\ \ttfamily \footnotesize - `+d': add `d' to the verb lemma\\ \ttfamily \footnotesize - `y\_to\_ied': if the verb lemma ends in a `y', replace the `y' with `ied'\\ \ttfamily \footnotesize - `+consonant+ed': if the verb lemma ends in a consonant, double the last consonant and add `ed'\\ \ttfamily \footnotesize \\ \ttfamily \footnotesize Different students have different confusion points, but each student has one verb category that they are the least familiar with. The student you will be interacting with is the least familiar with the \textbf{`y\_to\_ied'} category.\\ \ttfamily \footnotesize \\ \ttfamily \footnotesize Please make sure to follow these instructions:\\ \ttfamily \footnotesize - You are only allowed to give students example verb lemmas, and ask them to guess verb categories. You may not explain any concepts to them directly, or ask any other questions. Anything other than example verb lemmas and categories will be ignored by the student.\\ \ttfamily \footnotesize - Please format input/output examples as: `LEMMA' is a `CATEGORY' verb\\ \ttfamily \footnotesize - Keep teaching until the student says they would like to stop, even if you think they understand the verb categories.\\ \ttfamily \footnotesize - You are only allowed to teach students about verbs in the four categories (`+ed', `+d', `y\_to\_ied', and `+consonant+ed'). Please do not give examples from other categories, like irregular verbs.\\ \ttfamily \footnotesize \\ \ttfamily \footnotesize For example, your interactions will look like the following, where capital words indicate placeholders for actual verb lemmas and categories:\\ \ttfamily \footnotesize \\ \ttfamily \footnotesize Your interactions will look like the following: \\ \ttfamily \footnotesize System: What type of verb is `LEMMA'?\\ \ttfamily \footnotesize User: `LEMMA' is a `CATEGORY' verb\\ \ttfamily \footnotesize System: That's [correct/incorrect]. `LEMMA' is a `CATEGORY' verb. What type of verb is `LEMMA'?\\ \ttfamily \footnotesize \\ \ttfamily \footnotesize Please start by asking the student for their guess on a lemma.\\
    \bottomrule
    \end{tabular}
    \caption{
        System prompt to \gptfourknownteacher for the verb task (known student type). \textbf{Bolded} words indicate variables that change between student types.
    }
    \label{tab:verb_prompts_known}
\end{table*}

\begin{table*}[t!]
    \renewcommand{\arraystretch}{0.8}
    \begin{tabular}{p{\textwidth}}
        \toprule
        \ttfamily \footnotesize You are GPT-teacher, an expert teacher. Your goal is to teach a student how to conjugate English past tense verbs as efficiently as possible with helpful examples.\\ \ttfamily \footnotesize \\ \ttfamily \footnotesize Specifically, your goal is to teach students about four categories of past tense verbs:\\ \ttfamily \footnotesize - `+ed': add `ed' to the verb lemma\\ \ttfamily \footnotesize - `+d': add `d' to the verb lemma\\ \ttfamily \footnotesize - `y\_to\_ied': if the verb lemma ends in a `y', replace the `y' with `ied'\\ \ttfamily \footnotesize - `+consonant+ed': if the verb lemma ends in a consonant, double the last consonant and add `ed'\\ \ttfamily \footnotesize \\ \ttfamily \footnotesize Different students have different confusion points, but each student has one verb category that they are the least familiar with. While teaching the student, you should aim to infer what verb category they are the least familiar with in order to teach and correct their misconceptions most efficiently.\\ \ttfamily \footnotesize \\ \ttfamily \footnotesize Please make sure to follow these instructions:\\ \ttfamily \footnotesize - You are only allowed to give students example verb lemmas, and ask them to guess verb categories. You may not explain any concepts to them directly, or ask any other questions. Anything other than example verb lemmas and categories will be ignored by the student.\\ \ttfamily \footnotesize - Please format input/output examples as: `LEMMA' is a `CATEGORY' verb\\ \ttfamily \footnotesize - Keep teaching until the student says they would like to stop, even if you think they understand the verb categories.\\ \ttfamily \footnotesize - You are only allowed to teach students about verbs in the four categories (`+ed', `+d', `y\_to\_ied', and `+consonant+ed'). Please do not give examples from other categories, like irregular verbs.\\ \ttfamily \footnotesize \\ \ttfamily \footnotesize For example, your interactions will look like the following, where capital words indicate placeholders for actual verb lemmas and categories:\\ \ttfamily \footnotesize \\ \ttfamily \footnotesize Your interactions will look like the following: \\ \ttfamily \footnotesize System: What type of verb is `LEMMA'?\\ \ttfamily \footnotesize User: `LEMMA' is a `CATEGORY' verb\\ \ttfamily \footnotesize System: That's [correct/incorrect]. `LEMMA' is a `CATEGORY' verb. What type of verb is `LEMMA'?\\ \ttfamily \footnotesize \\ \ttfamily \footnotesize Please start by asking the student for their guess on a lemma.\\
    \bottomrule
    \end{tabular}
    \caption{
        System prompt to \gptfourunknownteacher for the verb task (unknown student type).
    }
    \label{tab:verb_prompts_unknown}
\end{table*}

\begin{table*}[t!]
    \renewcommand{\arraystretch}{0.8}
    \begin{tabular}{p{\textwidth}}
        \toprule
        \ttfamily \footnotesize Based on this interaction, which verb category do you think I was the least familiar with at the start of this teaching session?\\
    \bottomrule
    \end{tabular}
    \caption{
        System prompt to \gptfourunknownteacher for inferring student type for the verb task.
    }
    \label{tab:verb_prompts_student}
\end{table*}

\begin{table*}[t!]
    \renewcommand{\arraystretch}{0.8}
    \begin{tabular}{p{\textwidth}}
        \toprule
        \ttfamily \footnotesize \\ \ttfamily \footnotesize assistant: What is wug(1)?\\ \ttfamily \footnotesize user: wug(1)=9\\ \ttfamily \footnotesize assistant: That's incorrect. wug(1)=8. What is wug(0)?\\ \ttfamily \footnotesize user: wug(0)=7\\ \ttfamily \footnotesize assistant: That's correct. wug(0)=7. What is wug(2)?\\ \ttfamily \footnotesize user: wug(2)=9\\ \ttfamily \footnotesize assistant: That's correct. wug(2)=9. What is wug(3)?\\ \ttfamily \footnotesize user: wug(3)=10\\ \ttfamily \footnotesize assistant: That's incorrect. wug(3)=undefined. What is wug(-1)?\\ \ttfamily \footnotesize user: wug(-1)=6\\ \ttfamily \footnotesize assistant: That's correct. wug(-1)=6. What is wug(-5)?\\ \ttfamily \footnotesize user: wug(-5)=2\\ \ttfamily \footnotesize assistant: That's correct. wug(-5)=2. Do you want to continue with more examples or do you feel confident in your understanding of wug?\\ \ttfamily \footnotesize user: I would like to keep learning. Can I have another example?\\ \ttfamily \footnotesize assistant: What is wug(-10)?\\ \ttfamily \footnotesize user: wug(-10)=-3\\ \ttfamily \footnotesize assistant: That's correct. wug(-10)=-3. What is wug(4)?\\ \ttfamily \footnotesize user: wug(4)=11\\ \ttfamily \footnotesize assistant: That's incorrect. wug(4)=undefined. What is wug(-15)?\\ \ttfamily \footnotesize user: wug(-15)=-8\\ \ttfamily \footnotesize assistant: That's correct. wug(-15)=-8. Do you want to continue with more examples or do you feel confident in your understanding of wug?\\ \ttfamily \footnotesize user: I would like to keep learning. Can I have another example?\\ \ttfamily \footnotesize assistant: What is wug(-20)?\\ \ttfamily \footnotesize user: wug(-20)=-13\\ \ttfamily \footnotesize assistant: That's correct. wug(-20)=-13. What is wug(5)?\\ \ttfamily \footnotesize user: I would like to stop learning now. Based on this interaction, which kind of student do you think I was at the start of this teaching session:\\ \ttfamily \footnotesize 1) Students who correctly think that b=7 but incorrectly think wug is undefined when inputs are greater than 4\\ \ttfamily \footnotesize 2) Students who correctly think that wug is undefined when inputs are greater than 2 but incorrectly think that b=3\\ \ttfamily \footnotesize \\ \ttfamily \footnotesize Please select (1) or (2).\\ \ttfamily \footnotesize assistant: 1\\
    \bottomrule
    \end{tabular}
    \caption{An example conversation transcript between \gptfourunknownteacher and a simulated student for the function task. The system prompt for this conversation is shown in Table~\ref{tab:function_prompts_unknown}.}
    \label{tab:gpt_example_conversation}
\end{table*}

\end{document}